\documentclass[poms,nonblindrev]{poms1} 

\OneAndAHalfSpacedXI 

\usepackage{graphicx}
\usepackage{subfigure, epsfig}
\usepackage{natbib}

\usepackage{mathtools,algorithm,algpseudocode,multicol,booktabs,multirow,ulem,hyperref,amsmath, color} 
\DeclareMathOperator{\sign}{sgn}

\bibpunct[, ]{(}{)}{,}{a}{}{,}%
\def\bibfont{\small}%
\def\newblock{\ }%
\newcommand{\HL}{\color{black}}

\TheoremsNumberedThrough     
\ECRepeatTheorems

\EquationsNumberedThrough    

\begin{document}
	
	
	\RUNAUTHOR{xxx}

	\RUNTITLE{Sharpe Ratio MDPs}
	
	\TITLE{Sharpe Ratio Optimization in Markov Decision Processes}
	 \ARTICLEAUTHORS{%
	 	\AUTHOR{Shuai Ma}
	 	\AFF{School of Business, Sun Yat-sen University, Guangzhou, China, \EMAIL{mash35@mail.sysu.edu.cn}} 
         \AUTHOR{Guangwu Liu}
	 	\AFF{College of Business, City University of Hong Kong, Hong Kong, China, \EMAIL{msgw.liu@cityu.edu.hk}}
	 \AUTHOR{Li Xia\footnotemark[1]}
	 \footnotetext[1]{Corresponding Author.}
	 \AFF{School of Business, Sun Yat-sen University, Guangzhou, China, \EMAIL{xiali5@sysu.edu.cn}}
 } 

\ABSTRACT{%
Sharpe ratio (also known as reward-to-variability ratio) is a widely-used metric in finance, which measures the additional return at the cost of per unit of increased risk (standard deviation of return). 
However, the optimization of Sharpe ratio in Markov decision processes (MDPs) is challenging, because there exist two difficulties hindering the application of dynamic programming. {\HL One is that dynamic programming does not work for fractional objectives, and the other is that dynamic programming is invalid for risk metrics}. 
In this paper, we study the Sharpe ratio optimization in infinite-horizon MDPs, considering both the long-run average and discounted settings. We address the first challenge with the Dinkelbach’s transform, which converts the Sharpe ratio objective to a mean-squared-variance (M2V) objective. It is shown that the M2V optimization and the original Sharpe ratio optimization share the same optimal policy when the risk-sensitive parameter is equal to the optimal Sharpe ratio. 
For the second challenge, we develop an iterative algorithm to solve the M2V optimization which is similar to a mean-variance optimization in MDPs. We iteratively solve the M2V problem and obtain the associated Sharpe ratio that is used to update the risk-sensitive parameter in the next iteration of M2V problems. We show that such a sequence of Sharpe ratios derived is monotonically increasing and converges to the optimal Sharpe ratio. For both average and discounted MDP settings, we develop a policy iteration procedure and prove its convergence to the optimum. Numerical experiments are conducted for validation. To the best of our knowledge, our approach is the first that solves the Sharpe ratio optimization in MDPs with dynamic programming type algorithms. We believe that the proposed algorithm can shed light on solving MDPs with other fractional objectives.
}%


\maketitle

%



\section{Introduction}

Decision-makers must balance risks and rewards in risk-sensitive situations.
In general, most of the risk measures $\phi$ can be regarded as a function of $ \eta $ and $ \rho $, where $ \eta $ refers to the expectation of a stochastic reward and $ \rho $ quantifies its uncertainty.
One of the most prevalent risk-sensitive optimization frameworks is mean-variance optimization~\citep{Markowitz1952,li2000,basak2010}, where 
$\phi (\eta, \rho) = \eta - \beta \rho$ is a linear combination of $ \eta $ (mean) and $ \rho $ (variance) with a risk-sensitive parameter $\beta$.
Another type of risk-sensitive criteria is the family of reward-risk ratios, which is typically defined as $ \phi (\eta, \rho) = \eta/\rho $.
Compared to the mean-variance optimality criterion where the risk-sensitive parameter $\beta$ could be subjective, a reward-risk ratio is more straightforward: how much reward is credited per unit of risk.

Reward-risk ratio optimization has drawn substantial attention among academics and practitioners in the field of risk management, and numerous ratio criteria have been investigated.
\citet{Sharpe66} proposes a reward-risk ratio, which is well known as the Sharpe ratio or the reward-to-variability ratio, where the standard deviation of the return serves as the risk measure.
Sharpe ratio is a relative measure of risk, since it is unit-free and dimension-free, as the units or dimensions possessed by the underlying variable are washed out by the division.
During the same period, \citet{Treynor65} defines the Treynor ratio which adopts beta as a risk measure.
Beta, sometimes referred to as levered or equity beta, is a risk measure that expresses how well a portfolio beats the overall market. It is an indicator of systematic risk that cannot be reduced by diversification. 
\citet{Sortino94} present the Sortino ratio which chooses the downside standard deviation as the risk measure. It is convincing since the upside volatility is a bonus for the reward and should not be concerned in the risk measure. 
In addition to the above reward-risk ratios, other general ratio measures include the Omega ratio~\citep{Keating02}, represented as $ \mathbb{E}\{ (\Phi - \tau)_+ \} / \mathbb{E}\{ (\tau - \Phi)_+ \} $ with $\Phi$ the random return and $ \tau \in \mathbb{R} $ a target reward value. Similarly, \citet{Biglova04} formulate the Rachev ratio as the division of two CVaRs, i.e., $ \rho_{\alpha}	\{ \tau - \Phi \} / \rho_{\alpha'} \{ \Phi-\tau \} $, where $\rho_{\alpha}\{\cdot\}$ and $\rho_{\alpha'}\{\cdot\}$ are CVaR measures with probability levels $\alpha, \alpha' \in [0,1]$. Both the above Omega ratio and Rachev ratio indicate that a lower downside risk and a higher upside reward are favored. 
Other typical ratios include the stable tail adjusted return ratio~\citep{Martin03}, Farinelli-Tibiletti ratio~\citep{Farinelli03}, etc.
For more details on ratio optimization, see~\citep{Farinelli08,Ogryczak17,Sehgal21}.

Although ratio optimization has numerous applications in finance~\citep{Lorig16,Guasoni20,Chakrabarti21,Atmaca22}, most of them focus on static optimization scenarios and very limited research has been done in the field of Markov decision processes (MDPs) and reinforcement learning (RL).
\citet{Sobel85} studies the Sharpe ratio optimization problem in a unichain MDP via parametric analysis of a linear program with the same number of variables and one more constraint than the formulation of gain-rate optimization, which guarantees a local optimality.
\citet{Huang94} unify the maximum Sharpe-ratio, negative variance-with-bounded-mean, and mean-penalized-by-variance MDPs with one mathematical program.
\citet{Tamar2012policy} propose a framework for policy gradient algorithms to a variety of variance-related problems including Sharpe ratio optimization in RL and demonstrate the algorithms' convergence to local minima.
\citet{Ji21} develop bisection-based algorithmic frameworks for robust portfolio optimization models under Sharpe and Omega ratio criteria, where optimality is attained by iteratively refining convex programming subproblems. Furthermore, they establish a data-driven optimization paradigm for maximizing ambiguous reward-risk ratio measures, employing semi-infinite programming techniques to reformulate the epigraphic representation of these fractional objectives.
{\HL
Suttle et al. (2021) propose model-free RL algorithms to cost-aware MDPs that optimize the long-run average-reward-to-cost ratio ($\mathbb{E}(\Phi) / \mathbb{E}(\Phi')$, where $\Phi$ and $\Phi'$ refer to the returns with respect to a reward function and a cost funtion, respectively).}
The proposed algorithms adopt an actor-critic framework with linear function approximation and a two-timescale strategy based on relative value iteration Q-learning.
Furthermore, \citet{Suttle22} develop a policy gradient optimization scheme for Omega ratio maximization where the numerator and denominator are conditional expectations.
\citet{Liu17} consider a distributionally robust reward-risk ratio optimization with moment constraints. 
By approximating the semi-infinite constraints with entropic risk, they convert the problem to a nonlinear semi-infinite programming problem with Lagrange duality.
From the related works we can see that, the non-convex of Sharpe ratio fundamentally precludes existing optimization paradigms, mathematical programming or policy gradient techniques, from attaining global optimality.

This paper focuses on dynamic programming (DP) for Sharpe ratio optimization in both average and discounted MDPs, which considers the \textit{mean} and \textit{steady-state variance} together. 
In the literature, the steady-state variance \citep{Filar89} is defined as $\zeta = \lim_{T \rightarrow \infty}  \frac{1}{T} \mathbb{E} \{ \sum_{t=1}^{T} (r(S_t) - \eta_a)^2 \} $, which evaluates how far the immediate rewards $r(S_t)$ deviate from the long-run average $\eta_a = \lim_{T \rightarrow \infty} \frac{1}{T}\mathbb{E}\{\sum_{t=1}^{T} r(S_t) \} $ during the whole process. 
Sharpe ratio optimization has a close relation to mean-variance optimization. For variance minimization, \cite{Filar89} demonstrate the justifiability of steady-state variance in average MDPs with a simple example. They convert the problem to convex quadratic programs and show the existence of deterministic optimal strategies.
\cite{Sobel94} and \cite{Chung94} each independently examine the variance minimization problem in unichain MDPs with a mean performance constraint. The problem can be converted to a sequence of parametric linear programs with the help of the extant theory on quasiconcave minimization, and the necessary characteristics and Pareto optimality are investigated. \cite{Prashanth2013NIPS} propose actor-critic algorithms to estimate policy gradients of the accumulated returns' variance in discounted MDPs and the steady-state variance in average MDPs. They prove the asymptotic local convergences of the algorithms using the approach of ordinary differential equations. For the mean-variance optimization in average RL, \citet{Gosavi2014} presents a model-free approach similar to Q-learning. Though the approach is validated with a numerical experiment, its convergence analysis is absent. This gap has been filled by~\cite{XIA2016} in the MDP framework, where a policy iteration is proposed for variance minimization in average MDPs. With the aid of the sensitivity-based optimization theory~\citep{Cao2007}, the author derives a variance difference formula which quantifies the
difference between steady-state variances under any two policies. A local convergence of the proposed policy iteration is proved with the difference formula. This work is later extended to the mean-variance optimization in average MDPs~\citep{Xia2020}. \cite{Bisi2020ijcai} study the mean-variance optimization in discounted RL, where the steady-state variance is assessed to bound the limiting average variance. They develop a gradient-based
trust region policy optimization~\citep{schulman15} algorithm with a monotonic policy improvement. \cite{Zhang2021} focus on the mean-variance optimization in both discounted and average MDPs and develop a policy iteration as well as a gradient-based RL algorithm. In order to address the policy-dependent reward function, they reformulate variance using its Legendre-Fenchel dual with an augmented variable. 
In fact, this variance reformulation resembles the {\it pseudo	variances} defined by \cite{XIA2016,Xia2020} in which more in-depth studies on the variance and mean-variance optimization problems, along with local convergences in average MDPs, are presented.

Most of the above studies in stochastic dynamic scenarios rely on mathematical programming-based~\citep{Sobel85,Sobel94,Filar89,Chung94,Huang94,Liu17,Ji21} or gradient-based methods~\citep{Tamar2012policy,Prashanth2013NIPS,Bisi2020ijcai,Zhang2021,Suttle22}.  
However, gradient-based methods suffer from intrinsic deficiencies: slow convergence, high variance of gradient estimations, and sensitivity to step sizes~\citep{ZHAO2012}. 
Mathematical programming-based methods, on the other hand, usually have substantially extended variables and are not scalable to large-scale problems. In recent decades, a widely used framework for handling large-scale MDPs is RL (e.g., the game of Go has a state space size $ 3^{19^2} \approx 10^{172}$), which usually depends on DP and optimality equations. Therefore, developing DP type approaches is of practical significance for solving ratio optimization problems.

However, the traditional DP and Bellman optimality equation are developed for objectives of expectation of random rewards but do not work for objectives in a ratio form. One solution to this problem is linearization, which converts the ratio objective to a linear combination of its numerator and denominator with a risk-sensitive parameter. Details have been provided later in Section~\ref{section_3}.
{\HL The other issue is that dynamic programming is invalid for most, if not all, risk measures, such as variance and value at risk}. 
This issue hinders the direct application of DP principle, and new optimization methodologies for risk-sensitive MDPs are desirable.

{\HL 
The contributions of this paper are two-fold. 
	\begin{itemize}
		\item First, we prove the equivalence between the Sharpe
		ratio optimization and its linearized M2V optimization. 
		By using the Dinkelbach’s transform, we convert the Sharpe ratio objective to a mean-squared-variance
		(M2V) objective under three assumptions, and show that they share the same optimal policy when the risk-sensitive parameter $\beta$
		equals the optimal Sharpe ratio. Then we show the existence of optimal deterministic policies and
		prove a policy improvement rule to generate a better policy with larger Sharpe ratio through solving
		an M2V optimization problem.
		Thus, the Sharpe ratio optimization in MDPs is converted to a series of
		M2V optimization in MDPs. 
		\item Second, we propose policy iteration type algorithms, named SRPI and
		SRPI+, to efficiently solve the Sharpe ratio optimization in both average and discounted MDPs.
		In each iteration, we solve the M2V optimization problem
		with the current $\beta$ such that an improved policy is obtained with a better Sharpe ratio, then we
		update $\beta$ with the associated Sharpe ratio and iteratively derive a new M2V problem. Repeating
		this procedure, we prove that such a sequence of Sharpe ratios derived is monotonically increasing
		and converges to the optimal Sharpe ratio. This policy iteration procedure works for both average
		and discounted MDP settings, and we summarize it as a trilevel optimization algorithm termed
		SRPI. By analyzing the structure property of M2V optimization problems, we further develop a
		modified algorithm called SRPI+ which utilizes extra mechanisms for efficiency. 
	\end{itemize}	
	As far as we are aware, our paper is the first that develops DP
	type algorithms for the Sharpe ratio optimization in MDPs with a guaranteed global optimality.
	Thus, our approach has potentials to be further utilized to guide the development of temporal-difference RL algorithms for large-scale scenarios. Moreover, our work may shed light on studying
	other types of ratio optimization in MDPs, since a similar equivalence between ratio optimization
	and a sequence of mean-risk optimization may also hold in these new problem settings.}

The remainder of the paper is organized as follows. In Section~\ref{section2}, we give the problem formulation of Sharpe ratio optimization in both average and discounted MDPs. In Section~\ref{section_3}, we present the main results of this paper, including the policy iteration type algorithms and the convergence analysis. In Section~\ref{section4}, we conduct numerical experiments to demonstrate the effectiveness of our approach. Finally, Section~\ref{section5} concludes this paper.


\section{Problem formulations}\label{section2}
This section presents the formulations of Sharpe ratio optimization in both average and discounted MDPs.
We first give the requisite mathematical notations and then formulate the Sharpe ratio optimization with expectation and variance value functions.

\subsection{Markov decision processes}

This paper focuses on discrete-time infinite-horizon MDPs with finite state and action spaces.
An average MDP is denoted by
$ \mathcal{M}_a := \langle \mathcal{S},  \mathcal{A}, r, p \rangle $,
where
$ \mathcal{S} $ is a finite state space and $S_t \in \mathcal{S}$ represents the state at decision epoch $t \in \mathbb{N}^+ := \{1,2,\cdots\}$; 
$ \mathcal{A}(s)$ is the admissible action set for $s \in \mathcal{S}$, $ \mathcal{A} := \bigcup_{s \in \mathcal{S}}  \mathcal{A}(s)$ is a finite action space, and $A_t \in  \mathcal{A}$ represents the action chosen at $t$;
$r: \mathcal{S} \times  \mathcal{A} \rightarrow \mathbb{R}$ is a reward function;
$p(s' \mid s, a) = \mathbb{P}(S_{t+1}=s' \mid S_t = s, A_t = a)$ denotes the time-homogeneous transition probability for $s, s' \in \mathcal{S}$ and $a \in  \mathcal{A}(s)$. 
Similarly, a discounted MDP is denoted by
$ \mathcal{M}_c := \langle \mathcal{S},  \mathcal{A}, r, p, \mu, \alpha \rangle $, where 
$ \alpha \in (0,1)$ is a discount factor; $ \mu \in \Delta( \mathcal{S})$ is an initial state probability mass function and $ \Delta( \mathcal{S}) $ is the probability simplex over $ \mathcal{S}$.
With a slight abuse of notation, we also use $r$ and $\mu$ to denote an $ | \mathcal{S}| $-dimensional column vector and an $ | \mathcal{S}| $-dimensional row vector, respectively, where $| \mathcal{S}|$ is the cardinality of $\mathcal{S}$.

A policy prescribes how to choose actions sequentially.
Let $H_t \coloneqq (S_1, A_1, S_2, \cdots, S_t) $ denote a $t$-step history and $\mathcal{H}_t$ the set of such histories.
A general policy space concerned in MDPs is a randomized history-dependent policy space $ \mathcal{D}_h$, where $d := (d_1, d_2, \cdots, d_N) \in  \mathcal{D}_h$ for $N \in \mathbb{N}^+ \bigcup \{ +\infty\}$ and $d_t: \mathcal{H}_t \rightarrow \Delta ( \mathcal{A}(S_t))$. 
A policy is stationary when $d_t$ is independent of time and deterministic if it specifies an action for each state.
Let $ \mathcal{D}$ denote a stationary deterministic policy space where $ d: \mathcal{S} \rightarrow  \mathcal{A}$, $\forall d \in  \mathcal{D}$.
Since it has been proved later that there exists an optimal stationary deterministic policy to the Sharpe ratio optimizations in both average and discounted settings, we focus on $  \mathcal{D} $ only.
A policy $d \in  \mathcal{D}$ induces a time-homogeneous Markov reward process (MRP). 
We denote its transition probability by an $ |\mathcal{S}| $-by-$ |\mathcal{S}| $ matrix $ P^d $ with $ P^d(s, s') = p(s' \mid s, d(s)) $ and its reward function by an $ |\mathcal{S}| $-dimensional column vector $ r^d $ with $ r^d(s) = r(s, d(s)) $, for $ s, s' \in \mathcal{S} $.
We further denote its stationary distribution by an $ |\mathcal{S}| $-dimensional row vector $ \pi^d $ with $ \pi^d P^d = \pi^d $. 
When it is clear in the context, we omit the subscript ``$ d $'' for notational simplicity.

\subsection{Sharpe ratio optimization}
{\HL
Sharpe ratio is defined by the ratio of excess reward ($\eta - \tilde{\eta}$) to standard deviation of reward,
and we  optimize it in both average and discounted MDPs.}
The average setting concentrates on the steady-state performance which is suitable for continuing (cyclical) tasks in domains such as manufacturing, scheduling, and queueing.
Conversely, the discounted setting emerges in finance, where the time value of rewards is highlighted.  
Moreover, the discounted formulation preserves theoretical consistency with stochastic horizons governed by geometric termination probabilities (\textit{cf.} Chapter 5.3 in \citep{Puterman1994a}).
Next, we present the Sharpe ratio optimization in the two settings. 

\subsubsection{Average setting}
In an average setting, we focus on ergodic MDPs, whose average reward is denoted by
\begin{equation} \nonumber
	\eta_a = \eta^d_a \coloneqq \lim_{T \rightarrow \infty} \frac{1}{T} \mathbb{E}^d \left \{
	\sum_{t=1}^{T} r(S_t) \right \} = \lim_{T \rightarrow \infty} \frac{1}{T} \sum_{t=1}^{T}(P^{t-1} r) (s), \quad \forall s \in \mathcal{S},
\end{equation}
where $\mathbb{E}^d$ represents the expectation of the Markov chain under policy $d$.
Since $P$ is aperiodic for any $d \in D$, there exists a limiting matrix $P^* \coloneqq \lim_{T \rightarrow \infty} P^{T} $.
Furthermore, since $P$ is recurrent for any $d \in D$, we have $P^*= e \pi$, where $e$ is a column vector with 1’s.
The average reward can be rewritten as
\begin{equation} \nonumber
	\eta_a = (P^* r) (s) = \pi r, \quad \forall s \in \mathcal{S}.
\end{equation}
Similarly, the steady-state variance is denoted by 
\begin{equation} \nonumber
	\zeta_a = \zeta^d_a \coloneqq \lim_{T \rightarrow \infty}  \frac{1}{T} \mathbb{E}^d \{ \sum_{t=1}^{T} (r(S_t) - \eta_a)^2 \} = \lim_{T \rightarrow \infty} \frac{1}{T} \sum_{t=1}^{T}[P^{t-1} (r- \eta_a e )^2_\odot]  (s) = \pi (r- \eta_a e )^2_\odot, \quad \forall s \in \mathcal{S},
\end{equation}
{\HL
where $(r - \eta_a e)_{\odot}^2$ is the component-wise
	square of vector $(r - \eta_a e)$, i.e.,
	\begin{equation*}
		(r - \eta_a e)_{\odot}^2 := ((r(s_1,d(s_1)) - \eta_a)^2, \
		(r(s_2,d(s_2)) - \eta_a)^2, \ \dots, \ (r(s_{|\mathcal{S}|},d(s_{|\mathcal{S}|})) - \eta_a)^2 )^\mathrm{T}.
	\end{equation*}}
The variable $\zeta_a$ quantifies the dispersion of immediate rewards from its average in the long run. 
We define the Sharpe ratio in an average setting by
\begin{equation} \nonumber
	\psi_a = \psi^d_a \coloneqq \frac{\eta_a - {\tilde \eta}_a}{\sqrt{\zeta_a}}, \quad \forall d \in  \mathcal{D}',
\end{equation}
where $ \mathcal{D}' = \{ d \in  \mathcal{D} \mid \zeta^d_a \neq 0 \}$. 
The largest risk-free reward ${\tilde \eta}_a  = \max_{d \in  \mathcal{D}} \{ \eta^d_a \mid \zeta^d_a = 0 \}$ if there exists at least one policy $d$ such that $\zeta^d_a = 0$, otherwise we set ${\tilde \eta}_a = 0$.

\subsubsection{Discounted setting}
In a discounted setting, the ergodicity assumption is ignored.
The normalized discounted mean value function under $ d \in  \mathcal{D} $ is denoted by 
\begin{equation}
	\label{discMean1}
	v(s) = v^{d}(s) \coloneqq (1 - \alpha) \mathbb{E}^d_s \left \{ \sum_{t=1}^{\infty} \alpha^{t-1} r(S_t)  \right \}, \quad \forall s \in \mathcal{S},
\end{equation}
where $ \mathbb{E}^d_s $ stands for the expectation given initial state $ S_1  = s $ under $ d $. 
We have the following matrix form
\begin{equation} \nonumber
	v = (1 - \alpha) (I - \alpha P)^{-1} r.
\end{equation}
Considering the initial state distribution $ \mu $, we have the normalized discounted mean as
\begin{equation}
	\label{discMean}
	\eta_c = \eta^d_c \coloneqq (1 - \alpha) \mathbb{E}^d_\mu \left\{ \sum_{t=1}^{\infty} \alpha^{t-1} r(S_t) \right \}   = \mu v,
\end{equation}
where $ \mathbb{E}^d_\mu $ stands for the expectation given $ S_1 \sim \mu $ under $ d $.
Similarly, the normalized discounted steady-state variance value function under $ d $ is denoted by
\begin{equation}
	\label{discVar1}
	w(s) = w^{d}(s) \coloneqq (1 - \alpha) \mathbb{E}^d_s \left \{\sum_{t=1}^{\infty} \alpha^{t-1} [r(S_t) - \eta_c]^2 \right \}, \quad \forall s \in \mathcal{S},
\end{equation}
and in matrix form, we have
\begin{equation} \nonumber
	\label{discVar1InMat}
	w = (1 - \alpha) (I - \alpha P)^{-1} (r - \eta_c e )^2_\odot
\end{equation}
and the normalized discounted steady-state variance
\begin{equation} \nonumber
	\label{discVar}
	\zeta_c = \zeta^d_c \coloneqq (1 - \alpha) \mathbb{E}^d_\mu \left\{ \sum_{t=1}^{\infty} \alpha^{t-1} [r(S_t) - \eta_c]^2 \right \}   = \mu w,
\end{equation}
which quantifies the cumulative discounted reward deviations from the normalized discounted mean.
Analogous to the stationary distribution $ \pi $ in an average setting, denote the normalized occupation measure $ \pi_c \coloneqq (1 - \alpha) \mu (I - \alpha P)^{-1} $, we have 
\begin{equation*}
\eta_c = \pi_c r, \qquad \zeta_c = \pi_c (r- \eta_c e )^2_\odot.
\end{equation*}
We define the Sharpe ratio in a discounted setting by
\begin{equation} \nonumber
	\psi_c = \psi^d_c \coloneqq \frac{\eta_c - {\tilde \eta}_c}{\sqrt{\zeta_c}}, \quad \forall d \in  \mathcal{D}',
\end{equation}
where ${\tilde \eta}_c  = \max_{d \in  \mathcal{D}} \{ \eta^d_c \mid \zeta^d_c = 0 \}$.
Next, we define the Sharpe ratio optimizations in the two settings.

\subsubsection{Sharpe ratio optimization in MDPs}


In the two settings above, we study the Sharpe ratio optimization defined by 
\begin{flalign}\label{eq:SRoriginal}
	& \mbox{(P0): \hspace{4cm} }
	\begin{array}{cc}
		&\psi^* = \max\limits_{d \in  \mathcal{D}'} \left\{ \frac{\eta^d - {\tilde \eta}}{\sqrt{\zeta^d}} \right\}, \\
		&d^* \in \argmax\limits_{d \in  \mathcal{D}'}\left\{ \frac{\eta^d - {\tilde \eta}}{\sqrt{\zeta^d}} \right\},
	\end{array}&
\end{flalign}
where $ (\psi, \eta, \zeta, {\tilde \eta}) \in \{ (\psi_a, \eta_a, \zeta_a, {\tilde \eta}_a), (\psi_c, \eta_c, \zeta_c, {\tilde \eta}_c) \} $.
To simplify the model, we remove the largest risk-free reward $\tilde{\eta}$ from the reward function if it exists.
\begin{assumption}
	\label{assumption:risk_free}
	The reward function in an MDP accounts for the excess reward.  
\end{assumption}
It should be noted that the Sharpe ratio maximization is equivalent to the variation coefficient minimization under Assumption~\ref{assumption:risk_free}. 
The \textit{coefficient of variation}, as a unitless statistical measure of relative dispersion, provides critical insights for comparative risk analysis in financial instruments, particularly when assessing cross-sectional heterogeneity in security portfolios~\citep{Miller77}. 
Its analytical utility extends beyond traditional risk assessment to meta-optimization frameworks, where it enables adaptive regulation between exploration and exploitation mechanisms in stochastic optimization algorithms, as demonstrated in the confidence-bound optimization paradigm~\citep{Jiang19}.
\begin{remark}[Opitmality equivalence to coefficient of variation]\label{remark:eq}	
    If $\eta^d \neq 0$ holds for all $d \in  \mathcal{D}$, the Sharpe ratio maximization is equivalent to the variation coefficient minimization, \textit{i.e.}, $ \max_{d \in  \mathcal{D}} \frac{\eta^d}{\sqrt{\zeta^d}} \Longleftrightarrow \min_{d \in  \mathcal{D}} \frac{\sqrt{\zeta^d}}{\eta^d} $.
\end{remark}

To ensure the legitimacy of the ratio, we keep the variances of risk-free policies strictly positive.
\begin{assumption}
	\label{assumption:positive_variance}
	 We set $ \zeta^d = M \in \mathbb{R}^+, \forall d \in \{ d' \in  \mathcal{D} \mid \zeta^{d'} = 0 \}$, where $M$ is a large number.
\end{assumption}
A large $M$ prevents policies with zero variance from being selected.
Under Assumptions~\ref{assumption:risk_free} and~\ref{assumption:positive_variance}, the Sharpe ratio in the two settings is $\psi \coloneqq \frac{\eta}{\sqrt{\zeta}}$,
where $ (\psi, \eta, \zeta) \in \{ (\psi_a, \eta_a, \zeta_a), (\psi_c, \eta_c, \zeta_c) \} $. 
This objective $\psi$ cannot be optimized directly with DP-based algorithms such as policy iteration or value iteration. To solve it in MDPs, we square the standard deviation for variance first.
It is usually assumed that $ \eta \geq 0$ holds for any $ d \in  \mathcal{D} $~\citep{Sobel85}. 
In many cases, this condition could be too strict, and we present an alternative as follows. 
\begin{assumption}
	\label{assumption:square}
	The optimal Sharpe ratio $ \psi^* \coloneqq \max_{d' \in  \mathcal{D}} \{ \psi^{d'} \}  > -\psi^d, \quad \forall d \in  \mathcal{D} $. 
\end{assumption}
This assumption ensures that $ \psi^* > 0 $ and $\psi^* > |\psi^d|$ hold for any $ d \in \{ d' \in  \mathcal{D} \mid \psi^{d'} < 0 \}$.
Under Assumptions~\ref{assumption:risk_free}-\ref{assumption:square}, the Sharpe ratio optimization (P0) is equivalent to the optimization 
\begin{flalign}\label{eq:SRMDP}
	&\mbox{(P1): \hspace{4cm} }
	\begin{array}{cc}
		&{\psi^*}^2 \coloneqq \max\limits_{d \in  \mathcal{D}} \left\{{\psi^d}^2 \right\} = \max\limits_{d \in  \mathcal{D}} \left\{ \frac{{\eta^d}^2}{\zeta^d} \right\}, \\
		&d^* \in \mathop{\arg \max}\limits_{d \in  \mathcal{D}} \left\{{\psi^d}^2 \right\},
	\end{array}&
\end{flalign}
where $ (\psi, \eta, \zeta) \in \{ (\psi_a, \eta_a, \zeta_a), (\psi_c, \eta_c, \zeta_c) \} $.

This paper studies the Sharpe ratio optimization (P1) in MDPs.
To the best of our knowledge, this problem remains unresolved within the DP framework. We bridge the gap with DP-based algorithms for (P1) while preserving the Bellman optimality principle, thereby addressing this theoretical challenge in risk-sensitive MDPs.

\section{Sharpe ratio dynamic programming}\label{section_3}
In this section, we solve the Sharpe ratio optimization (P1) with DP in average and discounted MDPs.
{\HL 
The failure of DP in the Sharpe ratio optimization is caused by two fundamental challenges, and our methodology addresses them as follows.
}
\begin{enumerate}
	\item Fractional objective: The ratio-form objective violates the additive structure required by DP's recursive optimality principle.
    We resolve it with the Dinkelbach's transform, which recasts the fractional objective into a sequence of linearized objectives. This iterative procedure maintains convergence guarantees while enabling Bellman operator decomposition.
	\item Variance optimization with DP: Direct optimization of variance introduces non-Markovian dependency that breaks standard DP recursions. We introduce a pseudo mean to preserve the Bellman equation's structural integrity and present DP-based global optimization algorithms to solve the M2V objectives resulted from linearization. 
\end{enumerate}

DP-based algorithms, such as policy iteration and value iteration, can therefore be developed for the Sharpe ratio optimization in both average and discounted MDPs.
Taking policy iteration for example, we propose a Sharpe ratio policy iteration (SRPI) algorithm and a modified version termed SRPI+ with better efficiency. 

\subsection{Dinkelbach's transform}
Ratio-form optimization problems constitute a well-established subclass of fractional programming (FP).
It is theoretically challenging to solve FP problems since they are nonconvex, which renders conventional convex optimization techniques inapplicable. Although structural guarantees exist for concave-convex FP, where quasi-convexity emerges from the numerator's concavity and denominator's convexity, general FP solutions require specialized treatment.
The Charnes-Cooper transform~\citep{Charnes62} reformulates such problems through auxiliary variable substitutions, but imposes affine equality constraints that violate the Bellman equation's structural prerequisites in DP contexts. This methodological incompatibility motivates our exclusive focus on Dinkelbach’s transform~\citep{Dinkelbach67}.
By geometrically characterizing this transform through a four-region operational taxonomy in the (denominator, numerator)-plane, we restate the Dinkelbach’s transform to generic ratio optimization with straightforward proofs and analyses.
This transform enables sequential linearization with strict positivity of the denominator---a condition naturally satisfied in Sharpe ratio contexts due to variance nonnegativity.

For two functions $f: \mathcal X \rightarrow \mathbb{R}$ and $g: X \rightarrow \mathbb{R}\backslash\{0\}$, a generic ratio optimization is defined as 
\begin{flalign}\label{eq:ratioOpt}
	& \mbox{(R0): \hspace{4cm} }
	\kappa^* = \max_{x \in \mathcal X} \kappa(x) \coloneqq \frac{f(x)}{g(x)}.&
\end{flalign}


\begin{theorem}[Ratio optimization equivalence]	\label{th:eq}
Given the optimal ratio $\kappa^*$, a solution $ x^* \in \mathcal X $ is optimal to the generic ratio optimization (R0) if and only if 
	\begin{equation}
		\label{eq:problemEq}
		x^* \in \mathop{\arg \max}_{x \in \mathcal X}  \{\sign(g(x)) (f(x) - \kappa^* g(x)) \},
	\end{equation} 
	where the sign function $ \sign(x) := \frac{x}{|x|}$ with $x \neq 0$.
\end{theorem}
\proof{Proof of Theorem~\ref{th:eq}.} 


\begin{align} 
		\max_{x \in \mathcal X} \left\{ \frac{f(x)}{g(x)} \right\}
		& \Leftrightarrow \max_{x \in \mathcal X} \left\{ \frac{f(x)}{g(x)} - \kappa^* \right\} \nonumber \\
		& = \max_{x \in \mathcal X} \left \{ \frac{f(x) - \kappa^*g(x) }{g(x)} \right \} \nonumber \\
		& = \max_{x \in \mathcal X} \left \{ \sign(g(x)) \frac{f(x) - \kappa^*g(x) }{|g(x)|} \right \} \nonumber \\
		& \Leftrightarrow \max_{x \in \mathcal X} \left\{ \sign(g(x)) (f(x) - \kappa^* g(x))  \right\}. \nonumber 
\end{align}

Since $ \max_{x \in \mathcal X} \left\{ \frac{f(x)}{g(x)} - \kappa^* \right\} = 0 \Leftrightarrow \max_{x \in \mathcal X} \left\{ \sign(g(x)) (f(x) - \kappa^* g(x))\right\} = 0  $, we have the last equivalence. 
Moreover, the equivalence holds for a minimization objective as well, {\it i.e.}, $\min_{x \in \mathcal X} \left\{ \frac{f(x)}{g(x)} \right\} \Leftrightarrow \min_{x \in \mathcal X} \left\{ \sign(g(x)) (f(x) - \kappa^* g(x)) \right\}$ with $\kappa^*$ the minimal ratio here. {\hfill \Halmos}
\endproof

Theorem~\ref{th:eq} reveals that a ratio optimization can be addressed by solving its linearized surrogate given $\kappa^*$.
Our adaptation of Dinkelbach's transform relaxes the original requirement by only necessitating denominator nonzero for operational validity. Two critical observations emerge from this transform.
\begin{enumerate}
	\item Optimal ratio dependency: The equivalence between ratio optimization and linearized optimization is conditional upon attaining the optimal ratio $\kappa^*$. However, it is an endogenous product rather than an exogenous input and cannot be prespecified in practice. Fortunately, it has been shown that given the sign of $g$ fixed, $\kappa^*$ can be asymptotically attained through successive optimizations of $\max_{x \in \mathcal X} \left\{ \sign(g(x)) (f(x) - \kappa g(x)) \right \}$ initialized from any $ \kappa \in \mathbb{R} $, 
with $\kappa$ updated to the ratio of an optimal policy to a preceding problem~\citep{Dinkelbach67}.
	\item Sign-sensitive optimization: Denominator sign definiteness (assumed positive in classical Dinkelbach's transform) fundamentally governs optimization geometry. When the sign of denominator is uncertain, the problem results in two different optimization problems, which are illustrated in the four quadrants of the ({\it g-f}) plane separately.
\end{enumerate}

\begin{figure}[!h]
	\centering	                                    
		\includegraphics[width=.6\textwidth, trim=5cm 3cm 14cm 4cm, clip]{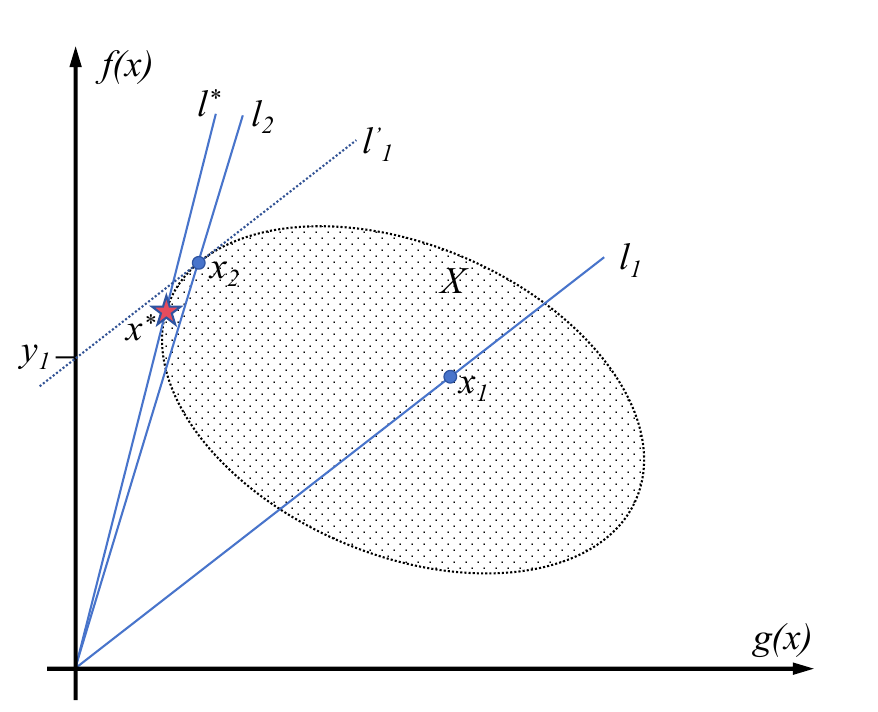}     
		\caption{The iterative process in the Dinkelbach's transform for the case where $f(x) \geq 0$ and $g(x) > 0$ (in the first quadrant, Q1).} \label{fig:Q1}
\end{figure}

\begin{figure}[hbtp]
	\centering
	\subfigure[Q2]
	{
		\includegraphics[width=.3\textwidth, trim=5.5cm .5cm 15cm 3.5cm, clip]{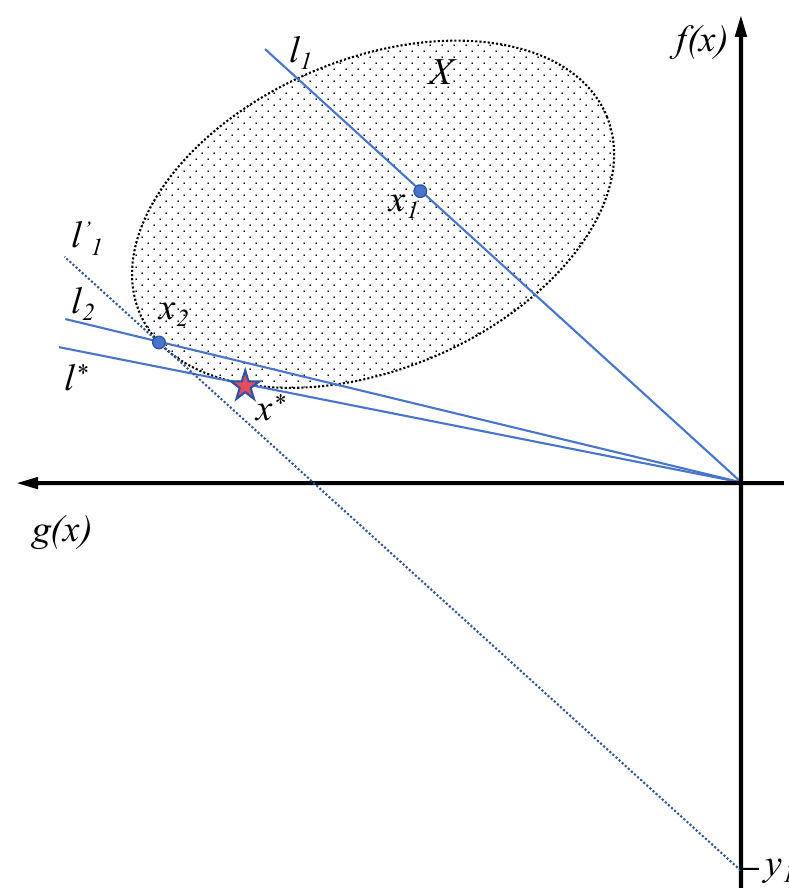} 
	}
	\subfigure[Q3]
	{
		\includegraphics[width=.3\textwidth, trim=7cm 3cm 14cm 3.5cm, clip]{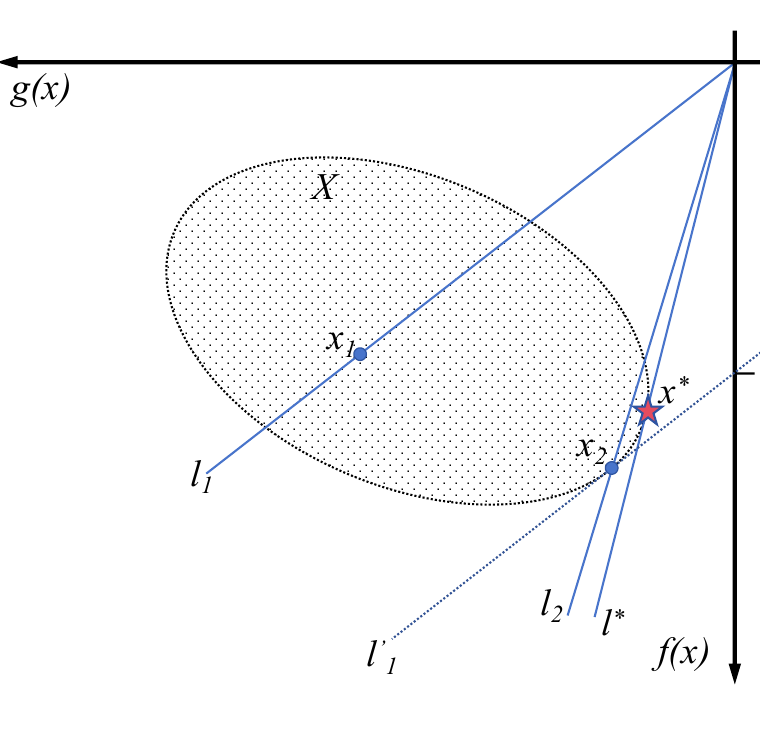} 
	}
	\subfigure[Q4]
	{
		\includegraphics[width=.3\textwidth, trim=5.5cm .5cm 14cm 3.5cm, clip]{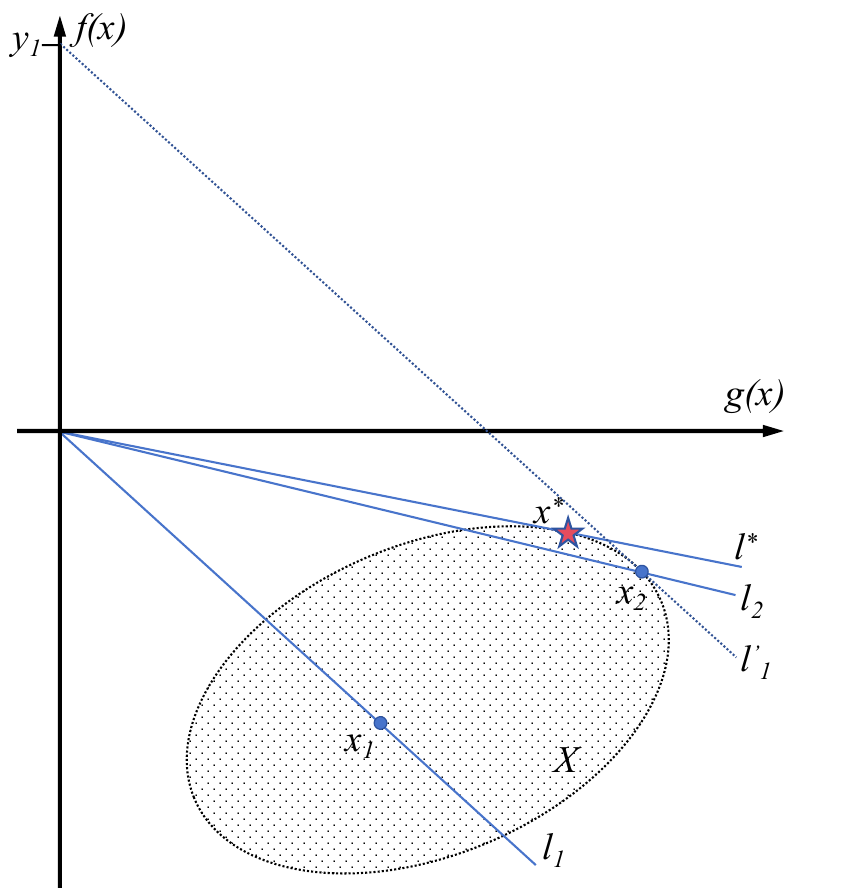} 
	}
	\caption{Cases in the second, third and fourth quadrants.} \label{fig:Q234}
\end{figure}

We illustrate the second point first in the case where $f(x) \geq 0$ and $g(x) > 0$ (the first quadrant) in Figure~\ref{fig:Q1}.
For any initial point $x_1 \in \mathcal X$, denote its ratio (slope of $l_1$) as $\kappa(l_1)$, the linearized optimization $\max_{x \in \mathcal X} \left\{ f(x) - \kappa(l_1) g(x) \right \}$ results in $l'_1$, and the optimal value equals the intercept $y_1$.
This optimization can be considered as moving $l_1$ with the slope $\kappa(l_1)$ as higher as possible (to maximize $y_1$), and its solution $x_2 \in\mathop{\arg\max}_{x \in \mathcal X} \{ f(x) - \kappa (l_1) g(x) \}$ can guarantee an improved slope $\kappa(l_2)$.
By repeating the process, it has been shown that $ \lim_{i \rightarrow \infty} x_i = x^{*}$, converging to the optimal ratio~\citep{Dinkelbach67}.
Similar iterative optimization processes are illustrated in the other three quadrants in Figure~\ref{fig:Q234}. 


\begin{corollary}[Analyses in four quadrants] 
	\label{cor:sign}
	Given $g(x) > 0$, a solution $ x^* \in  \mathcal{X} $ is optimal to the ratio optimization (R0) if and only if 
	\begin{equation}
		\label{eq:positiveDenom}
		x^* \in \mathop{\arg \max}_{x \in \mathcal{X}}  \{ (f(x) - \kappa^* g(x)) \}.
	\end{equation} 
	While given $g(x) < 0$,	we have 
	\begin{equation}
		\label{eq:negativeDenom}
		x^* \in \mathop{\arg \min}_{x \in \mathcal{X}}  \{ (f(x) - \kappa^* g(x)) \}.
	\end{equation}
\end{corollary}
\proof{Proof of Corollary~\ref{cor:sign}.} 
For the case Q1 where $f(x) \geq 0$ and $g(x) > 0$, we have the problem
\begin{flalign}\label{eq:Q1}
	& \mbox{(R1): \hspace{4cm} }
	\max_{x \in \mathcal X} \left\{ f(x) - \kappa^* g(x) \right\}.&
\end{flalign}
For the case Q2 where $f(x) \geq 0$ and $g(x) < 0$, we have the problem
\begin{flalign*}
	&\mbox{(R2): \hspace{4cm} }
	\begin{array}{ll}
		\max\limits_{x \in \mathcal X} \left\{ \frac{f(x)}{g(x)} \right\} & = \max\limits_{x \in \mathcal X} \left\{ -\frac{f(x)}{-g(x)} \right\}  \\
		&= -\min\limits_{x \in \mathcal X} \left\{ \frac{f(x)}{|g(x)|} \right\}  \\ 
        &\Leftrightarrow -\min\limits_{x \in \mathcal X} \left \{ f(x) - (-\kappa^*)|g(x)|  \right \} \\ 
        &= \max\limits_{x \in \mathcal X} \left \{ -f(x) + \kappa^*g(x) \right \}. 
	\end{array}&
\end{flalign*}
For the second equation, the equivalence holds as stated in Theorem~\ref{th:eq}.
Note that since there exists a minus before the optimization, its optimal value is $-\kappa^*$ instead of $\kappa^*$. 
Finally, since $\mathop{\arg \max}_{x \in \mathcal X} \left \{ -f(x) + \kappa^*g(x) \right \} = \mathop{\arg \min}_{x \in \mathcal X}  \{ (f(x) - \kappa^* g(x)) \}$, Corollary~\ref{cor:sign} is proved in the second quadrant.

For the case Q3 where $f(x) \leq 0$ and $g(x) < 0$, we have the problem
\begin{align} 
	\max_{x \in \mathcal X} \left\{ \frac{f(x)}{g(x)} \right\} 
	& = \max_{x \in \mathcal X} \left\{ \frac{-f(x)}{-g(x)} \right\} \nonumber \\
	& = \max_{x \in \mathcal X} \left\{ \frac{|f(x)|}{|g(x)|} \right\} \nonumber \\
	& \Leftrightarrow \max_{x \in \mathcal X} \left \{ |f(x)| - \kappa^*|g(x)|  \right \} \nonumber  \\
	& = \max_{x \in \mathcal X} \left \{ -f(x) + \kappa^*g(x) \right \}. \nonumber
\end{align}
For the case Q4 where $f(x) \leq 0$ and $g(x) > 0$, we have the problem
\begin{align} 
	\max_{x \in \mathcal X} \left\{ \frac{f(x)}{g(x)} \right\} 
	& = \max_{x \in \mathcal X} \left\{ -\frac{-f(x)}{g(x)} \right\} \nonumber \\
	& = -\min_{x \in \mathcal X} \left\{ \frac{|f(x)|}{g(x)} \right\}  \nonumber \\
	& \Leftrightarrow -\min_{x \in \mathcal X} \left \{ |f(x)| - (-\kappa^*)g(x)  \right \}  \nonumber \\
	& = \max_{x \in \mathcal X} \left \{ f(x) - \kappa^*g(x) \right \}. \nonumber
\end{align} {\hfill \Halmos}
\endproof

Corollary~\ref{cor:sign} illustrates that 
the optimization problem (R1) arises from the cases Q1 and Q4, while the optimization problem (R2) arises from the cases Q2 and Q3.
This structural dichotomy underlies the transform's reliance on denominator positivity. 
Both problems can be further unified to the optimization problem defined in (\ref{eq:problemEq}) and resonate with Theorem~\ref{th:eq}.
In short, the Dinkelbach's transform admits extension to the three cases if the standard assumptions do not hold.
\begin{enumerate}
	\item When the denominator is negative, the transform can be applied since $ \max_{x \in \mathcal X} \left\{ \frac{f(x)}{g(x)} \right\} = -\min_{x \in \mathcal X} \left\{ \frac{f(x)}{-g(x)} \right\} $, {\it i.e.}, the transform can be applied when the sign of denominator is deterministic;
	\item When the sign of denominator is uncertain, but the numerator is nonzero and its sign is deterministic, the transform can be applied since $ \max_{x \in \mathcal X} \left\{ \frac{f(x)}{g(x)} \right\} = \min_{x \in \mathcal X} \left\{ \frac{g(x)}{f(x)} \right\} $, {\it i.e.}, the transform can be applied when the sign of either numerator or denominator is deterministic;
	\item When the sign of either numerator or denominator is uncertain, the transform results in the two sign-constrained subproblems. 
\end{enumerate}

With the transform,  an iterative algorithm can be developed for generic ratio optimization  where a sequence of linearized surrogate optimizations is solved until the optimal ratio is attained.
Next, we propose an improvement rule for the ratio optimization problem in the following theorem.
\begin{theorem}[Improvement in the iterative scheme]
	\label{th:policyImp}
	For any $ \kappa \leq \kappa^* $, the optimal policy 
	\begin{equation}
		\label{eq:policyImp}
		x' \in \mathop{\arg \max}_{x \in \mathcal X} \{\sign(g(x)) (f(x) - \kappa g(x)) \}
	\end{equation} 
	ensures that $ \frac{f(x')}{g(x')} \geq \kappa $, and the equality holds if and only if $ \kappa = \kappa^* $.
	
\end{theorem}
\proof{Proof of Theorem~\ref{th:policyImp}.} 
We have $\sign(g(x')) (f(x') - \kappa g(x'))  = \max_{x \in \mathcal X} \{ \sign(g(x)) (f(x) - \kappa g(x)) \} \geq \sign(g(x^*))(f(x^*) - \kappa g(x^*)) \geq 0 $, 
where the last inequality holds since 
\begin{align} 
	\sign \left\{\sign(g(x^*))(f(x^*) - \kappa g(x^*))\right\}
	& = \sign \left\{\frac{\sign(g(x^*))(f(x^*) - \kappa g(x^*))}{|g(x^*)|}\right\} \nonumber \\
	& = \sign \left\{\frac{f(x^*) - \kappa g(x^*)}{g(x^*)}\right\}  \nonumber \\
	& = \sign \left\{ \kappa^* - \kappa \right\} \nonumber
\end{align}
and $ \kappa^* \geq \kappa $. Moreover, the last equality holds only when $ \kappa = \kappa^* $. 
Therefore, with $\sign(g(x')) (f(x') - \kappa g(x')) \geq 0$, we have $\frac{f(x')}{g(x')} \geq \kappa$ and the equality holds if and only if $\kappa = \kappa^*$.
{\hfill \Halmos}
\endproof

Theorem~\ref{th:policyImp} shows that once $ \kappa < \kappa^* $, the linearized optimization (\ref{eq:policyImp}) yields policies with strictly improved ratios.
It generates a monotonically increasing sequence of ratios that converge to the optimum, and the policy improvement continues until $ \kappa = \kappa^* $.
Moreover, a trivial case is that for any initial $ \kappa > \kappa^* $, it goes back to the case $ \kappa \leq \kappa^* $ after the first iteration.
Consequently, starting from any $ \kappa \in \mathbb{R} $, the generic ratio optimization (R0) can be solved with a sequence of linearized optimizations, provided the sign of $g$ remains prerequisite.
For a ratio optimization with $g: X \rightarrow \mathbb{R}^+$, the Dinkelbach's transform reduces to iteratively solving linearized problems R($\kappa$) in Algorithm~\ref{alg:ratioGlobal}.
\begin{flalign*}
	& \mbox{(R($\kappa$)): \hspace{4cm} }
	\max_{x \in \mathcal X} \left\{  (f(x) - \kappa g(x)) \right\}.&
\end{flalign*}

%
%


\begin{algorithm}[H]
	\caption{Dinkelbach's transform to ratio optimization}
	\label{alg:ratioGlobal}
	\begin{algorithmic} [1]
		
		\Require
		Initialize two parameters $ \kappa, \kappa' \in \mathbb{R} $ with $ \kappa \neq \kappa' $
		\Ensure
		An optimal solution and the optimal ratio
		
		\While {$ \kappa \neq \kappa' $}
		\State $ \kappa \leftarrow \kappa' $
		\State Solve the linearized problem R($\kappa$) and obtain an optimal solution $ x $
		\State $ \kappa' \leftarrow \frac{f(x)}{g(x)} $
		
		\EndWhile
		\State Return $ x $ and $ \kappa $
	\end{algorithmic}
\end{algorithm}

%

{\HL 
The convergence rate of Dinkelbach's transform can be studied under mild conditions as follows.
	\begin{itemize}
		\item The feasible space $\mathcal X$ is compact;
		\item functions $f$ and $g$ is continuous on $\mathcal X$;
		\item the denominator $g(x) \geq C > 0$ for all $x \in \mathcal X$ where $C$ is a constant; and
		\item the optimal solution $x^*$ is unique.
	\end{itemize}	
	Under the mild assumptions above, we give the following theorem.
	\begin{theorem}[Superlinear convergence rate]
		\label{th:conRate}
		The convergence rate of Dinkelbach's transform is superlinear.		
	\end{theorem}
	\proof{Proof of Theorem~\ref{th:conRate}.} 
		We define $f_t \coloneqq f(x_t)$, $g_t \coloneqq g(x_t)$, $f^* \coloneqq f(x^*)$ and $g^* \coloneqq g(x^*)$ for notational simplicity. 
		For the sequence $\{ \kappa_t \}$ where $\kappa_{t+1} \coloneqq f(x_t)/g(x_t) = f_t/g_t$, we prove the superlinear convergence rate by proving that
		\begin{equation}
			\label{eq:conRate}
			\lim\limits_{t \rightarrow \infty} \frac{|\kappa_{t+1} - \kappa^*|}{|\kappa_t - \kappa^*|} = 0.
		\end{equation}
		Without loss of generality, let us start the Dinkelbach's transform from $\kappa_0 = 0$.
		Consider the $t$-th iteration where
		\begin{equation*}
			f_t - \kappa_t g_t \geq f(x) - \kappa_t g(x), \quad \forall x \in \mathcal X,
		\end{equation*}
		and set $x = x^*$ and we have 
		\begin{align*} 
			f_t - \kappa_t g_t & \geq f(x^*) - \kappa_t g(x^*) \\
			& = \kappa^* g^* - \kappa_t g^* \\
			& = g^* (\kappa^* - \kappa_t) \\
			f_t & \geq g^* (\kappa^* - \kappa_t) + \kappa_t g_t \\
			\kappa^* g_t - f_t & \leq \kappa^* g_t - g^* (\kappa^* - \kappa_t) - \kappa_t g_t \\
			& = (g_t - g^*) (\kappa^* - \kappa_t) \\
			\frac{\kappa^* g_t - f_t}{g_t} & \leq \frac{g_t - g^*}{g_t} (\kappa^* - \kappa_t) \\
			\frac{|\kappa_{t+1} - \kappa^*|}{|\kappa_t - \kappa^*|} & \leq \frac{g_t - g^*}{g_t}.
		\end{align*}
		Since $x^*$ is unique, we have $\lim\limits_{t \rightarrow \infty} g_t - g^* = 0$ and Equation~\ref{eq:conRate} holds. {\hfill \Halmos}
\endproof
}

Thus far, we have addressed the first challenge posed at the beginning of this section by reformulating the ratio optimization into a sequence of linearized optimizations. 
However, for Sharpe ratio optimization (and most risk-sensitive ratio optimization problems) in MDPs, the linearized optimizations prove fundamentally incompatible with DP since the DP principle and Bellman optimality equation do not hold for MDPs with variance-related criteria (and most other risk measures). 
Next, we develop a DP-based global optimization algorithm to address this challenge. 

\subsection{Mean-squared-variance optimization}
{\HL 
An important research stream in multiperiod mean-variance optimization adopts a stochastic control perspective. The seminal works of \citep{li2000, Zhou00} reformulate the mean-variance portfolio selection problem as a linear-quadratic (LQ) control problem, employing an embedding technique to derive an elegant analytical solution through an iterative procedure. In this framework, the existence of a closed-form solution \citep{li2000} combined with the strict convexity of the efficient frontier \citep{Markowitz1952} establishes the equivalence between Sharpe ratio optimization and mean-variance optimization for standard multiperiod portfolio problems. However, this approach relies on an LQ model with linear state transitions that, while well suited for portfolio selection, offers limited generalization compared to Markov decision processes. Consequently, the methodology cannot be extended to general MDP settings, as the closed-form solution is specific to the particular structure of the portfolio selection problem.
}

We solve the Sharpe ratio optimization (P1) with a sequence of M2V optimization problems.
First, we show that (P1) is equivalent to an optimization in which only the variance item cannot be optimized with DP directly. 
Denote a random variable $ Q_a $ with support $ \{ r^d(s) \}_{s \in \mathcal{S}} $ and $ \mathbb{P}(Q_a = z) = \sum_{s \in \{ s' \in \mathcal{S}: r(s')=z \} } \pi (s) $ in an average setting, and $ Q_c $ with support $ \{ r^d(s) \}_{s \in \mathcal{S}} $ and $ \mathbb{P}(Q_c = z) = \sum_{s \in \{ s' \in \mathcal{S}: r(s')=z \} } \pi_c (s) $ in a discounted setting. 
{\HL 
With the random variables introduced, we have $\zeta = \mathbb{E}_\mu \{Q^2\}-\eta^2$, where $(Q, \eta, \zeta) \in \{ (Q_a, \eta_a, \zeta_a), (Q_c, \eta_c, \zeta_c) \}$.}
Letting $\kappa = 1 + {\psi}^2$ for notational simplicity, we have the following theorem.

\begin{theorem}[Sharpe ratio optimization equivalence]\label{th:OptEq}
	In both average and discounted MDPs, the mean-squared-variance (M2V) optimization
    \begin{flalign*}
	& \mbox{$(\mathcal{M}(\kappa^*))$: \hspace{2cm} }
	\max\limits_{d \in  \mathcal{D}} \left\{\mathbb{E}^d_\mu \left\{ Q^2 \right\} - \kappa^* \zeta^d \right\}, \quad (Q, \psi, \zeta) \in \{ (Q_a, \psi_a, \zeta_a), (Q_c, \psi_c, \zeta_c) \}&
    \end{flalign*}
	shares the same optimal policy space with the Sharpe ratio optimization (P1).
\end{theorem}
\proof{Proof of Theorem~\ref{th:OptEq}.} 
	The Sharpe ratio optimization (P1) in both average and discounted MDPs has the following problem equivalence.
	\begin{align*} 
		\max\limits_{d \in  \mathcal{D}} \left\{ \frac{{\eta^d}^2}{\zeta^d} \right\} & \Longleftrightarrow \max\limits_{d \in  \mathcal{D}} \left\{{\eta^d}^2 - {\psi^*}^2 \zeta^d \right\} \\
		& = \max\limits_{d \in  \mathcal{D}} \left\{\mathbb{E}^d_\mu \left\{ Q^2 \right\} - (1 + {\psi^*}^2) \zeta^d \right\} \\
		& = \max\limits_{d \in  \mathcal{D}} \left\{\mathbb{E}^d_\mu \left\{ Q^2 \right\} - \kappa^* \zeta^d \right\}.
	\end{align*}
	The first equivalence is from Corollary~\ref{cor:sign}, and the second holds since $ \eta^2 = \mathbb{E}_\mu \left\{ Q^2 \right\} - \zeta$, $\forall (Q, \eta, \zeta) \in \{ (Q_a, \eta_a, \zeta_a),  (Q_c, \eta_c, \zeta_c) \}$. {\hfill \Halmos}
\endproof

\begin{remark}
	We can apply the linearization after the conversion of $\eta^2$ as well, \textit{i.e.}, 
	\begin{align*} 
		{\psi^*}^2 & = \max\limits_{d \in  \mathcal{D}} \left\{ \frac{{\eta^d}^2}{\zeta^d} \right\} \\ 
		& = \max\limits_{d \in  \mathcal{D}} \left\{ \frac{\mathbb{E}^d_\mu \left\{ Q^2 \right\} - \zeta^d}{\zeta^d} \right\} \\
		{\psi^*}^2 + 1 & = \max\limits_{d \in  \mathcal{D}} \left\{ \frac{\mathbb{E}^d_\mu \left\{ Q^2 \right\}}{\zeta^d} \right\},
	\end{align*}
	which results in the same optimization $\mathcal{M}(\kappa^*)$.
\end{remark}

While $\kappa^*$ is unknown \textit{a priori}, Theorem~\ref{th:policyImp} enables solving (P1) with a sequence of M2V optimizations ($\{ \mathcal{M}(\kappa) \}$) with $\mathcal{M}(\kappa)$ defined by
\begin{flalign}\label{eq:M2VOpt3}
	& \mbox{($\mathcal{M}(\kappa)$): \hspace{4cm} }
	\max\limits_{d \in  \mathcal{D}} \left\{\mathbb{E}^d_\mu \left\{ Q^2 \right\} - \kappa \zeta^d \right\}.&
\end{flalign}
Next, we demonstrate the existence of deterministic optimal policies for $\mathcal{M}(\kappa)$ as well as (P1).

\begin{theorem}[Optimal deterministic policy to Sharpe ratio] \label{lemma:optDet}
	There exists a deterministic optimal policy to the Sharpe ratio optimization (P1) as well as any linearized optimization $\mathcal{M}(\kappa)$ with $\kappa \geq 0$ in both average and discounted MDPs.
\end{theorem}
\proof{Proof of Theorem~\ref{lemma:optDet}.} 
{\HL
Taking the average setting for example, we formulate an M2V problem $\mathcal{M}(\kappa^*)$ with a dual linear program as follows.
	\begin{equation*}
		\max\limits_{x} \left\{ \kappa^* \left [\sum\limits_{s \in
			\mathcal S}\sum\limits_{a \in \mathcal A(s)} r(s,a)x(s,a)\right ]^2 +(1 - \kappa^*)\sum\limits_{s \in
			\mathcal S}\sum\limits_{a \in \mathcal A(s)} r^2(s,a) x(s,a)
		\right\} 
	\end{equation*}
	\begin{equation*} \label{eq_LP}
		\begin{array}{cl}
			\mbox{s.t., }& \sum\limits_{a \in \mathcal A(s)}x(s,a) =
			\sum\limits_{s' \in \mathcal S}\sum\limits_{a \in
				\mathcal A(s')} p(s|s',a) x(s',a), \qquad \forall s \in \mathcal S, \\
			&\sum\limits_{s \in S}\sum\limits_{a \in \mathcal A (s)} x(s,a) = 1, \\
			&x(s,a) \geq 0, \qquad \forall s \in \mathcal S,  a \in \mathcal A (s).
		\end{array}
	\end{equation*}
	The objective is derived from
	\begin{equation*}
		\max\limits_{x} \left\{ \sum\limits_{s \in
			\mathcal S}\sum\limits_{a \in \mathcal A(s)} r^2(s,a) x(s,a) - \kappa^*\sum\limits_{s \in
			\mathcal S}\sum\limits_{a \in \mathcal A(s)} r^2(s,a) x(s,a) + \kappa^* \left [\sum\limits_{s \in
			\mathcal S}\sum\limits_{a \in \mathcal A(s)} r(s,a)x(s,a)\right ]^2
		\right\}.
	\end{equation*}
	Since the objective is convex on $x$ and the feasible region which is a bounded polyhedron (also known as a polytope), the maximum occurs at an extreme point which corresponds to a deterministic optimal policy.
	The conclusion works for both average and discounted settings.
	See (Huang and Kallenberg 1994) and Theorem 8.9.10 in (Puterman 1994) for details.
	Since Theorem 3 reveals that the Sharpe ratio optimization (P1) is equivalent to $\mathcal{M}(\kappa^*)$ with the risk-sensitive parameter $\kappa^* \geq 0$, the Sharpe
	ratio optimization has optimal deterministic policies in both average and discounted MDPs.}
{\hfill \Halmos}
\endproof

\begin{figure}[h]
	\centering
	\includegraphics[width=0.8\columnwidth]{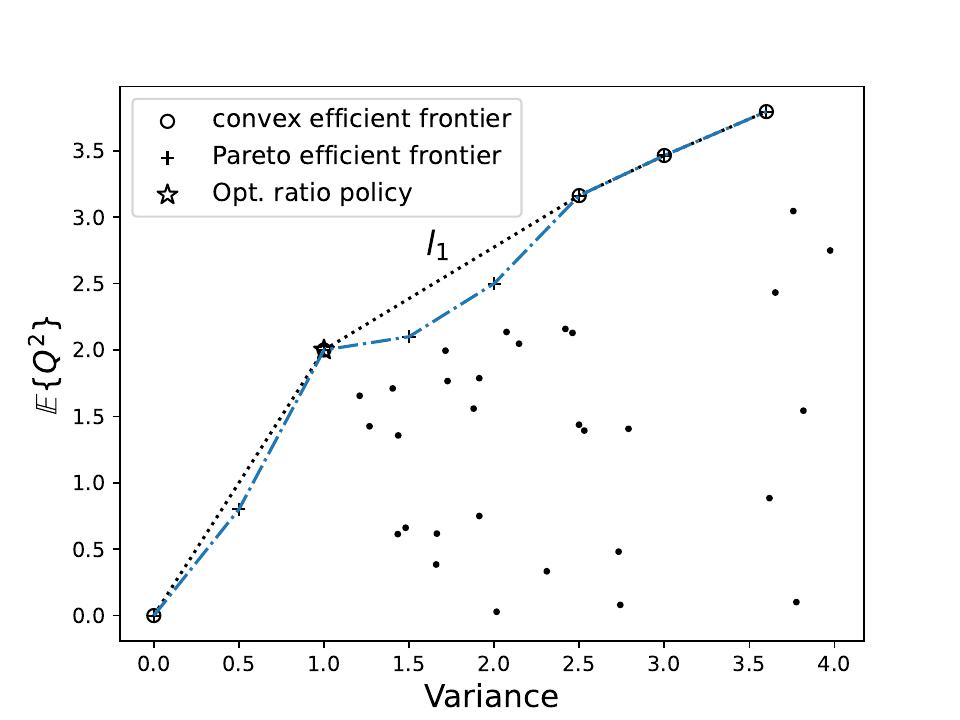}    
	\caption{The convex efficient frontier in the variance-mean-squared plane.}  \label{fig:tangents}
\end{figure}

Now we can limit the policy space to the stationary deterministic policy space $  \mathcal{D} $.
In Figure~\ref{fig:tangents}, all deterministic policies are represented with concrete dots in the variance-mean-squared plane. 
For the ratio and related M2V optimizations, a crucial frontier is defined as follows.
\begin{definition}[Convex efficient frontier]
    For the set $  \mathcal{D} \bigcup \{ (0, 0) \}$ in the mean-variance-squared coordinate system, the geometric boundary of its convex hull and Pareto efficiency frontier intersect to be its convex efficient frontier.
\end{definition}

The convex efficient frontier illustrated in Figure~\ref{fig:tangents} is composed of the circles, which are the intersection of the edge of the convex hull (represented by the dashed line) and the Pareto frontier (represented by "$ + $").
This frontier serves as the fundamental solution space for the Sharpe ratio optimization in the following way.

\begin{corollary}[Convergence along convex efficient frontier] \label{cor:conv}
    When solving (P1) with Theorem~\ref{th:policyImp} with an initial $\kappa =0$, the optimal policy evolves from the top (with the maximal expectation) down to the one closet to the origin along the convex efficient frontier, and the Sharpe ratio increases monotonically.
\end{corollary}
It is intuitive to interpret Corollary~\ref{cor:conv} in Figure~\ref{fig:Q1}. 
Each $\mathcal{M}(\kappa)$ with $\kappa \in [0, \kappa^*)$ maximizes the vertical intercept $y_1$ and produces a policy with decreased expectation and variance as well as increased Sharpe ratio.
Moreover, Theorem~\ref{th:eq} can be improved for finite MDPs (which have finite state and action spaces) as follows. 

\begin{corollary}[Problem equivalence in deterministic policy space]
	\label{cor:eqSR}
	A policy $ d^* $ is optimal to the Sharpe ratio optimization (P1) if and only if 
	\begin{equation}
		\label{optPolicyCond2}
		d^* \in \mathop{\arg \max}_{d \in  \mathcal{D}} \{\mathbb{E}^d_\mu \{ Q^2 \} - \kappa \zeta^d \}, \quad \forall \kappa \in (\kappa', \kappa^*],
	\end{equation} 
	where $ \kappa' $ is the slope of the line determined by the optimal ratio policy and its neighbor along the frontier farther to the origin (\textit{i.e.}, $ \kappa' = \kappa (l_1)$).
\end{corollary}

{\HL Building upon recent advances on mean-variance optimization in~\citep{Xia23}, we can solve $\mathcal{M}(\kappa)$ by DP-based algorithms with guaranteed global optimality.
In both average and discounted settings ({\it cf.} Remark~5 in \citep{Xia23}), a linearized optimization $\mathcal{M}(\kappa)$ with $\kappa \in \mathbb{R}^+$ is
\begin{equation}
	\max_{d \in  \mathcal{D}} \{\mathbb{E}^d_\mu \{ Q^2 \} - \kappa \zeta^d \} = \max_{d \in  \mathcal{D}} \{\mathbb{E}^d_\mu \{ Q^2 \} - \kappa \mathbb{E}^d_\mu \{ (Q - \eta^d )^2 \} \},
\end{equation}
where $ (Q, \zeta, \eta) \in \{ (Q_a, \zeta_a, \eta_a),  (Q_c, \zeta_c, \eta_c) \} $.
Since $\eta^d$ depends on $d$, $\mathcal{M}(\kappa)$ cannot be solved by traditional DP.
To solve this problem, we introduce a pseudo mean $y \in \mathbb{R}$ to transform $\mathcal{M}(\kappa)$ to an auxiliary standard MDP $\mathcal{M}(\kappa, y)$ 
with a reshaped reward function 
\begin{equation}
	r'(s, a) \coloneqq r^2(s, a) - \kappa [r(s, a) - y]^2, \quad \forall s \in  \mathcal{S}, a \in  \mathcal{A}(s).
\end{equation}
A locally optimal policy $d(y)$ to $\mathcal{M}(\kappa)$ can be achieved by iteratively updating $y$~\citep{XIA2016}, and it \textit{dominates} any policy $d'$ with $\eta^{d'} \in [y-|y-\eta^{d(y)}|, y+|y-\eta^{d(y)}|]$, \textit{i.e.}, 
\begin{equation*}
	\mathbb{E}^{d'}_\mu \{ Q^2 \} - \kappa \zeta^{d'} \leq \mathbb{E}^{d(y)}_\mu \{ Q^2 \} - \kappa \zeta^{d(y)}.
\end{equation*}
Since $\eta^d \in [\underline{r}, \overline{r}]$ for any $d \in \mathcal D$, where $\underline{r} \coloneqq \min\limits_{s \in  \mathcal{S}, a \in  \mathcal{A}(s)} r(s, a)$ and $\overline{r} \coloneqq \max\limits_{s \in  \mathcal{S}, a \in  \mathcal{A}(s)} r(s, a)$, we can achieve a locally optimal policy set $\{ d(y) \}_{y \in [\underline{r}, \overline{r}]}$ that dominates the targeted interval $[\underline{r}, \overline{r}]$, \textit{i.e.}, 
\begin{equation*}
	[\underline{r}, \overline{r}] \subset \bigcup\limits_{y \in [\underline{r}, \overline{r}]} [y-|y-\eta^{d(y)}|, y+|y-\eta^{d(y)}|].
\end{equation*}
Since the Sharpe ratio of any policy in $\{ d(y) \}$ has been evaluated, we can identify a globally optimal policy within $\{ d(y) \}$.}
Next, we propose a global optimization algorithm for the Sharpe optimization (P1) with DP. 

\subsection{Sharpe ratio policy iteration}

Since $\mathcal{M}(\kappa, y)$ is a standard MDP and can be solved by DP, we can now solve the Sharpe ratio optimization (P1) with DP in both average and discounted settings.
Taking the policy iteration for example, we present a Sharpe ratio policy iteration (SRPI) in Algorithm~\ref{alg:SRPI0}, which solves (P1) by solving a sequence ($\mathcal{M} (\kappa)$), and each $\mathcal{M} (\kappa)$ is solved by solving a sequence ($\mathcal{M} (\kappa, y)$) with a standard policy iteration.  

\begin{algorithm}[h]
	\caption{Sharpe ratio policy iteration (SRPI)}
	\label{alg:SRPI0}
	\begin{algorithmic} [1]
		
		\Require
		The MDP $\mathcal{M}$; initialize two parameters $ \kappa, \kappa' \in \mathbb{R} $ with $ \kappa \neq \kappa' $
		
		\Ensure
		An optimal policy and the optimal ratio 
		
		
		\While {$ \kappa \neq \kappa' $}
		\State $ \kappa = \kappa' $, 
		$\mathbb Y = [\underline{r}, \overline{r}]$,
		$  \mathcal{D}_s = \emptyset$ \Comment{Solve $\mathcal{M} (\kappa)$}
		
		\While {$ \mathbb Y \neq \emptyset $} 
		\State Sort the intervals in $\mathbb Y$ as
		$\mathbb Y = \{\mathbb Y_1, \mathbb Y_2, \dots\}$
		\State $y =
		(\max\{\mathbb Y_1\} + \min\{\mathbb Y_1\})/2$
		
		\State Solve $\mathcal{M}(\kappa, y)$ with policy iteration and acquire a policy $d$ \Comment{Solve $\mathcal{M} (\kappa, y)$}
		
		\State $\mathbb Y = \mathbb Y - [y-|y-\eta^{d}|, y+|y-\eta^{d}|]$
        \State $ \mathcal{D}_s =  \mathcal{D}_s \bigcup \{d\}$
		
		\EndWhile
		\State $d^{\dagger} \in \mathop{\arg \max}\limits_{d \in  \mathcal{D}_s} \{\mathbb{E}^{d}_\mu \{ Q^2 \} - \kappa \zeta^{d} \}$, $\kappa' = \frac{\mathbb{E}^{d^{\dagger}}_\mu \{ Q^2 \}}{\zeta^{d^{\dagger}}}$
		\EndWhile
		\State Return $ d^{\dagger} $ and $ \kappa $
	\end{algorithmic}
\end{algorithm}

Since the convex efficient frontier consists of finite policies, we have the following corollary for the computational complexity of Algorithm~\ref{alg:SRPI0}.
\begin{corollary}\label{cor:complexity}
    The computational complexity of Algorithm~\ref{alg:SRPI0} is $(| \mathcal{D}|+1)*(2| \mathcal{D}|+1)$ times of solving a standard MDP.
\end{corollary}
The Corollary~\ref{cor:complexity} offers a loose upper bound of computational complexity.
With Corollary~\ref{cor:conv}, the maximal number of $\mathcal{M} (\kappa)$'s for the Sharpe ratio optimization is $(| \mathcal{D}|+1)$ since the number of policies in the convex efficient frontier is bounded by $| \mathcal{D}|$ and a final optimization on $\mathcal{M} (\kappa^*)$ is needed for optimality guarantee.
See Corollary 1 in \citep{Xia23} for the computational complexity of $\mathcal{M} (\kappa)$.

The computational efficiency can be improved with two mechanisms.
First, we may go on to solve a new $\mathcal{M} (\kappa')$ directly once a policy with $\kappa' > \kappa$ is achieved.
Second, a policy subset becomes dominated once a policy with a positive $\mathcal{M}(\kappa)$ value is identified. 
\begin{proposition}[Extra domination] \label{prop:domainShrinkage}
	For a policy $d^\dagger \in  \mathcal{D}$ with $\mathbb{E}^{d^\dagger}_\mu \{ Q^2 \} - \kappa \zeta^{d^\dagger} >0 $, any policy in the set $\left \{ d \in  \mathcal{D} \bigm| |\eta^d| \leq \sqrt{\frac{\mathbb{E}^{d^\dagger}_\mu \{ Q^2 \} - \kappa \zeta^{d^\dagger}}{\kappa}} \right \}$ is dominated by $d^\dagger$.
\end{proposition} 
\proof{Proof of Proposition \ref{prop:domainShrinkage}.}
For any $d \in  \mathcal{D} $ such that $ |\eta^d| \leq \sqrt{\frac{\mathbb{E}^{d^\dagger}_\mu \{ Q^2 \} - \kappa \zeta^{d^\dagger}}{\kappa}}$, we have $ \mathbb{E}^{d}_\mu \{ Q^2 \} - \kappa [ \mathbb{E}^{d}_\mu \{ Q^2 \} - {\eta^d}^2 ] = \kappa {\eta^d}^2 + (1 - \kappa) \mathbb{E}^{d}_\mu \{ Q^2 \} \leq \kappa {\eta^d}^2 \leq \mathbb{E}^{d^\dagger}_\mu \{ Q^2 \} - \kappa \zeta^{d^\dagger} $.
{\hfill \Halmos}
\endproof

We utilize the above two mechanisms and propose a new algorithm named SRPI+ (Algorithm~\ref{alg:SRPI2}).
It is worth noting that, the two mechanisms can be applied after solving either an $\mathcal{M}(\kappa)$ or an $\mathcal{M}(\kappa, y)$, and SRPI+ is only \textit{statistically} faster than SRPI since starting from a better policy cannot guarantee a faster convergence.

\begin{algorithm}[h]
	\caption{Modified Sharpe ratio policy iteration (SRPI+)}
	\label{alg:SRPI2}
	\begin{algorithmic} [1]
		
		\Require
		The MDP $\mathcal{M}$; initialize two parameters $ \kappa, \kappa' \in \mathbb{R} $ with $ \kappa \neq \kappa' $
		
		\Ensure
		An optimal policy and the optimal ratio 
		
		\While {$ \kappa \neq \kappa' $}
		
		\State $ \kappa = \kappa' $, $\mathbb Y = [\underline{r}, \overline{r}]$,
		$ \mathcal{D}_s = \emptyset$  \Comment{Solve $\mathcal{M} (\kappa)$}
		
		\While {$ \kappa \geq \kappa' $ \textbf{and} $ \mathbb Y \neq \emptyset $} 
		
		\State Sort the intervals in $\mathbb Y$ as
		$\mathbb Y = \{\mathbb Y_1, \mathbb Y_2, \dots\}$
		\State $y =
		(\max\{\mathbb Y_1\} + \min\{\mathbb Y_1\})/2$
		
		\State Solve $\mathcal{M}(\kappa, y)$ with policy iteration and acquire a policy $d$ \Comment{Solve $\mathcal{M} (\kappa, y)$}
        
		\State $\mathbb Y = \mathbb Y - \left[y-|y-\eta^{d}|, y+|y-\eta^{d}|\right] - \left[-\sqrt{\frac{\mathbb{E}^{d}_\mu \{ Q^2 \} - \kappa \zeta^{d}}{\kappa}}, \sqrt{\frac{\mathbb{E}^{d}_\mu \{ Q^2 \} - \kappa \zeta^{d}}{\kappa}} \right]$, $\kappa' = \frac{\mathbb{E}^{d}_\mu \{ Q^2 \}}{\zeta^{d}}$
        
		\State $ \mathcal{D}_s =  \mathcal{D}_s \bigcup \{d\}$
		
		
		
		
		\EndWhile
        
		\If{$ \mathbb Y = \emptyset $}

		\State $d^{\dagger} \in \mathop{\arg \max}\limits_{d \in  \mathcal{D}_s} \{\mathbb{E}^{d}_\mu \{ Q^2 \} - \kappa \zeta^{d} \}$ 
		\EndIf
		
		\EndWhile
		
		\State Return $ d^{\dagger} $ and $ \kappa $
	\end{algorithmic}
\end{algorithm}



{\HL
In this section, we deal with the failure of dynamic programming in the Sharpe ratio optimization.
We apply the Dinkelbach’s transform to the fractional form of objective under Assumptions~\ref{assumption:risk_free}-\ref{assumption:square}, and solve (P1) with a sequence of M2V problems.
We analyze the properties of M2V problem and solve it in a similar way as a mean-variance optimization problem, which is thoroughly studied in \citep{Xia23}.
}
We propose a trilevel DP-based algorithm SRPI for solving the Sharpe ratio optimization (P1).
The outer level solves the Sharpe ratio optimization with $\{ \mathcal{M}(\kappa) \}$.
To iteratively update $\kappa$, we solve $\mathcal{M}(\kappa)$ with a global optimization algorithm in the middle level of SRPI.
The global optimization algorithm solves $\mathcal{M}(\kappa)$ with $ \mathcal{D}$ fully dominated by the solutions to $ (\mathcal{M}(\kappa, y)) $. 
For any $y \in [\underline{r}, \overline{r}]$, the inner level of SRPI solves $\mathcal{M}(\kappa, y)$ with $d \in  \mathcal{D}$ which dominates a subspace $\{ d' \in  \mathcal{D} \mid \eta^{d'} \in [y-|y-\eta^{d}|, y+|y-\eta^{d}|] \} $.
Through this approach, we address (P1) via a sequence of M2V optimizations where the associated ratio sequence converges monotonically to the optimum in finite iterations, while the resultant optimal policy for the final linearized problem $\mathcal{M}(\kappa^*)$ simultaneously constitutes an optimal strategy to (P1).
Moreover, a modified version SRPI+ with a \textit{statistically} faster convergence is also proposed.

\section{Numerical experiment}\label{section4}

We consider a small-scale MDP with a state space $ \mathcal{S} = \{ s_1, s_2, s_3 \}$, each of which has an action space $ \mathcal{A} = \{a_1, a_2, a_3\}$. 
For $s_1$, we have $p(\cdot \mid s_1, a_1) = (0.2, 0.4, 0.4)$, $p(\cdot \mid s_1, a_2) = (0.4, 0.3, 0.3)$, $p(\cdot \mid s_1, a_3) = (0.3, 0.4, 0.3)$, and $r(s_1, \cdot) = (5, 9, 7)$; 
For $s_2$, we have $p(\cdot \mid s_2, a_1) = (0.3, 0.3, 0.4)$, $p(\cdot \mid s_2, a_2) = p(\cdot \mid s_2, a_3) = (0.4, 0.3, 0.3)$, and $r(s_2, \cdot) = (5, 2, 0)$; 
For $s_3$, we have $p(\cdot \mid s_3, a_1) = (0.3, 0.3, 0.4)$, $p(\cdot \mid s_3, a_2) = (0.4, 0.4, 0.2)$, $p(\cdot \mid s_3, a_3) = (0.3, 0.4, 0.3)$, and $r(s_3, \cdot) = (2, 4, 2)$. 
We solve the Sharpe ratio optimization (P1) in an average setting by the two SRPI algorithms with an initial policy $d_0 = (a_1, a_1, a_1)$.

First, we compare the optimization processes of SRPI and SRPI+ algorithms listed in Tables~\ref{tab:optProc0} and \ref{tab:optProc1}, respectively.
In this example, both of the two algorithms share the same intermediate $\mathcal{M}(\kappa)$'s at the outer iterations and stop at the third outer iteration. Both algorithms converge to the same optimal policy $(a_1, a_1, a_2)$ which is marked with superscript ``$^*$" since its Sharpe ratio is the largest so far with $\kappa = 99$. 
The M2V and $\kappa$ values of each optimal policy to $\mathcal{M}(\kappa, y)$ are recorded.
At the first two outer levels of SRPI ($\mathcal{M}(0)$ and $\mathcal{M}(8.8910)$), the target interval $\mathbb Y = [0, 9]$ must be fully covered, while the processes for SRPI+ end quickly since a policy with a larger $\kappa$ is achieved.
For example, though the extra domination mechanism is prohibited at the first outer iteration when $\kappa = 0$, SRPI+ jumps to the second outer iteration directly with an improved $\kappa$ ($8.8910 > 0$).
When $\kappa = \kappa^*$, the two mechanisms fail to work and a full coverage must be implemented to guarantee global optimality, and the two algorithms share the same process at the final outer iteration ($\kappa=99$).
In addition, the extra domination mechanism is unused at the second outer iteration of SRPI+ though it is activated.
It implies that when the two mechanisms are implemented at the same level, the first one may render the second useless.

\begin{table}[htbp]
	\centering
	\caption{\label{tab:optProc0}The convergence process of SRPI algorithm. The bolded row is for the converged optimal policy.}
	\begin{tabular}{p{1.5cm}p{1.5cm}p{7cm}p{2cm}p{2cm}}
		\toprule
		$\kappa$ & $y$ & Policies $d$ during iteration & M2V &  $\kappa'$ \\
		\midrule[1pt]
		\multirow{3}{=}{0} & \multirow{1}{=}{4.5} & {($a_1$, $a_1$, $a_1$), ($a_2$, $a_1$, $a_2$), ($a_2$, $a_1$, $a_2$)} & 42.8257 & 8.8910 \\
		& \multirow{1}{=}{7.5826} & {($a_2$, $a_1$, $a_2$), ($a_2$, $a_1$, $a_2$)} & 42.8257 & 8.8910\\
		& \multirow{1}{=}{1.4174} & {($a_2$, $a_1$, $a_2$), ($a_2$, $a_1$, $a_2$)} & 42.8257 & 8.8910\\
		\midrule[1pt]
		\multirow{9}{=}{8.8910} & \multirow{1}{=}{4.5} & {($a_1$, $a_1$, $a_1$), ($a_1$, $a_1$, $a_2$), ($a_1$, $a_1$, $a_2$)} & 20.0242 & 99 \\
		& \multirow{1}{=}{6.8333} & {($a_1$, $a_1$, $a_2$), ($a_3$, $a_1$, $a_2$), ($a_3$, $a_1$, $a_2$)}& 16.8293 & 20.0970  \\
		& \multirow{1}{=}{8.6556} & {($a_3$, $a_1$, $a_2$), ($a_2$, $a_1$, $a_2$), ($a_2$, $a_1$, $a_2$)}& 0 & 8.8910  \\
		& \multirow{1}{=}{5.0110} & {($a_2$, $a_1$, $a_2$), ($a_1$, $a_1$, $a_2$), ($a_1$, $a_1$, $a_2$)}& 20.0242 & 99  \\
		& \multirow{1}{=}{2.1667} & {($a_1$, $a_1$, $a_2$), ($a_1$, $a_2$, $a_1$), ($a_1$, $a_2$, $a_1$)}& -6.5336 & 5.4519  \\
		& \multirow{1}{=}{3.6208} & {($a_1$, $a_2$, $a_1$), ($a_1$, $a_1$, $a_2$), ($a_1$, $a_1$, $a_2$)}& 20.0242 & 99  \\
		& \multirow{1}{=}{0.7125} & {($a_1$, $a_1$, $a_2$), ($a_1$, $a_3$, $a_1$), ($a_1$, $a_3$, $a_1$)}& -26.3895 & 2.2680  \\
		\midrule[1pt]
		\multirow{8}{=}{99} & \multirow{1}{=}{4.5} & {($a_1$, $a_1$, $a_1$), ($a_1$, $a_1$, $a_2$), ($a_1$, $a_1$, $a_2$)} & 0 & 99  \\
		& \multirow{1}{=}{6.8333} & {($a_1$, $a_1$, $a_2$), ($a_3$, $a_1$, $a_2$), ($a_3$, $a_1$, $a_2$)} & -118.4974 & 20.0970 \\
		& \multirow{1}{=}{8.6556} & {($a_3$, $a_1$, $a_2$), ($a_2$, $a_1$, $a_2$), ($a_2$, $a_1$, $a_2$)} & -434.0342 & 8.8910 \\
		& \multirow{1}{=}{5.0110} & {($a_2$, $a_1$, $a_2$), ($a_1$, $a_1$, $a_2$), ($a_1$, $a_1$, $a_2$)} & 0 & 99 \\
		
		& \multirow{1}{=}{2.1667} & {($a_1$, $a_1$, $a_2$), ($a_1$, $a_2$, $a_1$), ($a_1$, $a_2$, $a_1$)} & -177.7264 & 5.4519 \\
		& \multirow{1}{=}{\bfseries{3.6208}} & \bfseries{($\mathbf{a_1}$, $\mathbf{a_2}$, $\mathbf{a_1}$), ($\mathbf{a_1}$, $\mathbf{a_1}$, $\mathbf{a_2}$), ($\mathbf{a_1}$, $\mathbf{a_1}$, $\mathbf{a_2}$)$^*$} & \bfseries{0} & \bfseries{99} \\
		& \multirow{1}{=}{0.7125} & {($a_1$, $a_1$, $a_2$), ($a_1$, $a_3$, $a_1$), ($a_1$, $a_3$, $a_1$)} & -385.4301 & 2.2680 \\
		\bottomrule[1pt]     
	\end{tabular}
\end{table}

\begin{table}[htbp]
	\centering
	\caption{\label{tab:optProc1}The convergence process of SRPI+ algorithm. The bolded row is for the converged optimal policy.}
	\begin{tabular}{p{1.5cm}p{1.5cm}p{7cm}p{2cm}p{2cm}}
		\toprule
		$\kappa$ & $y$ & Policies $d$ during iteration & M2V &  $\kappa'$ \\
		\midrule[1pt]
		\multirow{1}{=}{0} & \multirow{1}{=}{4.5} & {($a_1$, $a_1$, $a_1$), ($a_2$, $a_1$, $a_2$), ($a_2$, $a_1$, $a_2$)} & 42.8257 & 8.8910 \\
		\midrule[1pt]
		\multirow{1}{=}{8.8910} & \multirow{1}{=}{4.5} & {($a_1$, $a_1$, $a_1$), ($a_1$, $a_1$, $a_2$), ($a_1$, $a_1$, $a_2$)}  & 20.0242 & 99 \\
		\midrule[1pt]
		\multirow{8}{=}{99} & \multirow{1}{=}{4.5} & {($a_1$, $a_1$, $a_1$), ($a_1$, $a_1$, $a_2$), ($a_1$, $a_1$, $a_2$)} & 0 & 99  \\
		& \multirow{1}{=}{6.8333} & {($a_1$, $a_1$, $a_2$), ($a_3$, $a_1$, $a_2$), ($a_3$, $a_1$, $a_2$)} & -118.4974 & 20.0970 \\
		& \multirow{1}{=}{8.6556} & {($a_3$, $a_1$, $a_2$), ($a_2$, $a_1$, $a_2$), ($a_2$, $a_1$, $a_2$)} & -434.0342 & 8.8910 \\
		& \multirow{1}{=}{5.0110} & {($a_2$, $a_1$, $a_2$), ($a_1$, $a_1$, $a_2$), ($a_1$, $a_1$, $a_2$)} & 0 & 99 \\
		
		& \multirow{1}{=}{2.1667} & {($a_1$, $a_1$, $a_2$), ($a_1$, $a_2$, $a_1$), ($a_1$, $a_2$, $a_1$)} & -177.7264 & 5.4519 \\
		& \multirow{1}{=}{\bfseries{3.6208}} & \bfseries{($\mathbf{a_1}$, $\mathbf{a_2}$, $\mathbf{a_1}$), ($\mathbf{a_1}$, $\mathbf{a_1}$, $\mathbf{a_2}$), ($\mathbf{a_1}$, $\mathbf{a_1}$, $\mathbf{a_2}$)$^*$} & \bfseries{0} & \bfseries{99} \\
		& \multirow{1}{=}{0.7125} & {($a_1$, $a_1$, $a_2$), ($a_1$, $a_3$, $a_1$), ($a_1$, $a_3$, $a_1$)} & -385.4301 & 2.2680 \\
		\bottomrule[1pt]     
	\end{tabular}
\end{table}

\begin{figure}[!ht]
	\centering
	\includegraphics[width=0.9\columnwidth]{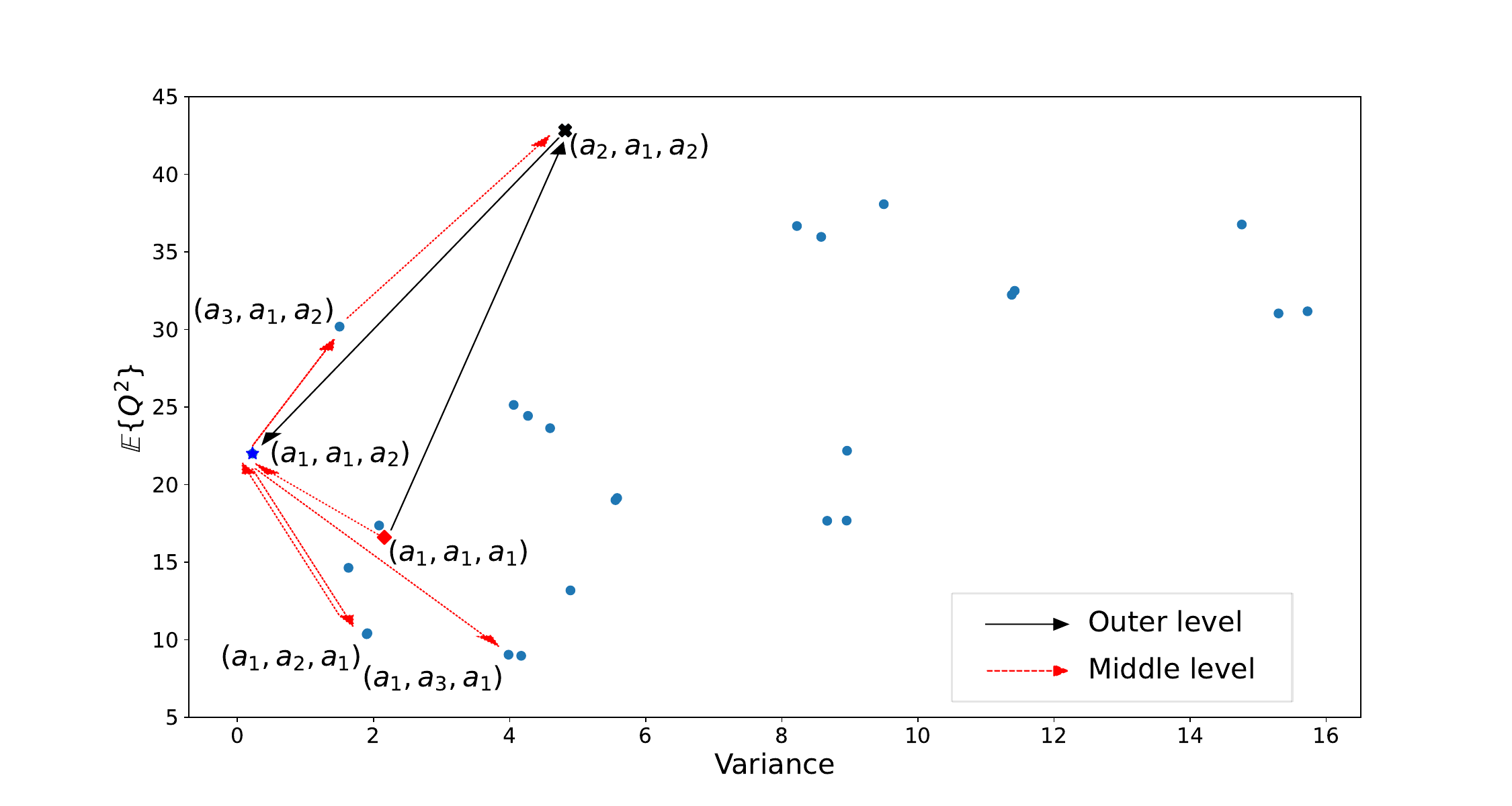} 
	\caption{The convergence processes of the two SRPI's at the outer levels (black solid arrows) and the middle levels at the final outer level (red dashed arrows). The optimal policies achieved at the outer level, ($a_2$, $a_1$, $a_2$) and ($a_1$, $a_1$, $a_2$), are on the convex efficient frontier, and ($a_1$, $a_1$, $a_2$), the closet policy to the origin along the frontier, is globally optimal.}     \label{fig:convProc}
\end{figure}

Next, we illustrate the optimization processes of the two algorithms in Figure~\ref{fig:convProc}.
Initializing from the policy marked by the red diamond ``$ \diamond $'', both algorithms solve $\mathcal{M}(8.8910)$ and $\mathcal{M}(99)$ sequentially at outer iterations, and the solutions are along the convex efficient frontier including ($a_2$, $a_1$, $a_2$), ($a_3$, $a_1$, $a_2$), and ($a_1$, $a_1$, $a_2$).
The optimal policy to $\mathcal{M}(8.8910)$ is marked by the black cross ``$ \times $'', and the optimal policy to $\mathcal{M}(99)$ marked by the blue star ``$ \star $'' is globally optimal to the Sharpe ratio optimization, since it is the one closet to the origin on the frontier.
During the optimization for $\mathcal{M}(8.8910)$ at the second outer level, both algorithms achieve the optimal policy ($a_1$, $a_1$, $a_2$) by solving the first standard MDP $\mathcal{M}(8.8910, 4.5)$, and SRPI+ updates $\kappa = \kappa'$ and goes on to solve $\mathcal{M}(8.8910, 99)$ due to the first mechanism, while SRPI has to cover $\mathbb Y$ by solving ($\mathcal{M}(8.8910, y)$).
To guarantee global optimality, full coverage is implemented at the final outer level in both algorithms, which is illustrated by the red dashed arrows.
Two points are worth noting. 
One resonating with Corollary~\ref{cor:conv} is that the policies visited at the outer level are always along the convex efficient frontier.
The other point is that some policies visited at the middle level are not on the frontier, such as the points ($a_1$, $a_2$, $a_1$) and ($a_1$, $a_3$, $a_1$).
This phenomenon is formally characterized by the quadratic distortion on the frontier ({\it cf.} Lemma 1 in \citep{XIA2016}).

\begin{figure}[!ht]
	\centering
	\includegraphics[width=0.9\columnwidth]{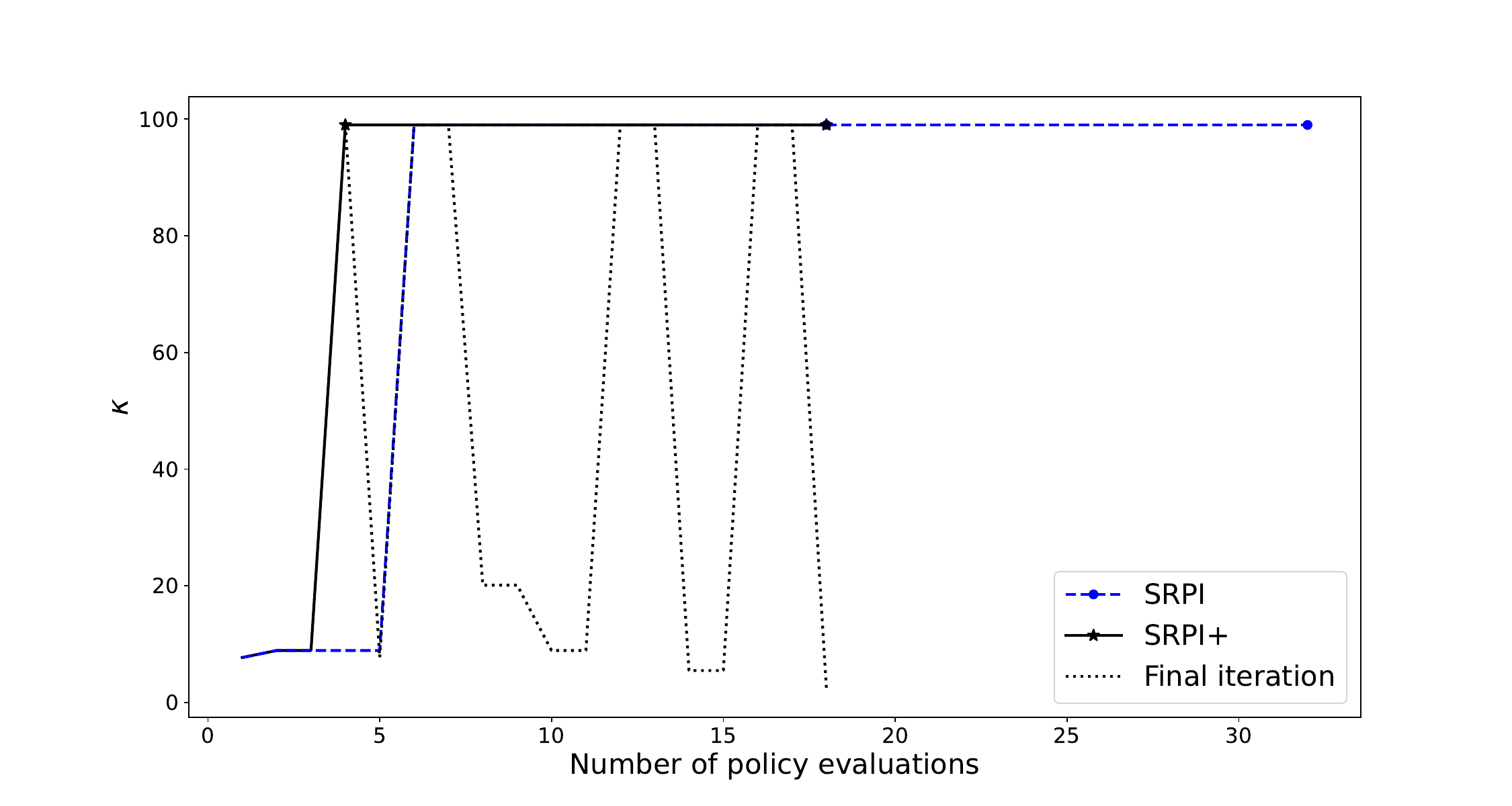} 
	\caption{A comparison on best-so-far $\kappa$'s between SRPI and SRPI+ algorithms.
	}     \label{fig:convCurve}
\end{figure}

Figure~\ref{fig:convCurve} compares the best-so-far $\kappa$'s along with the number of policy evaluations.
The final outer iterations of the two SRPIs are marked by dot ``$ \bullet $'' for SRPI and star ``$ \star $'' for SRPI+.
Note that given the same initial policy, the convergence processes of final outer iterations (between two identical markers) of the two algorithms are identical since $\mathcal{M}(\kappa^*)$ is unique. 
A complete coverage of $\mathbb Y$ must be implemented in solving $\mathcal{M}(\kappa^*)$ to guarantee optimality.
Moreover, the black dashed curve records $\kappa$ values during the final iteration in SRPI+,  
which are identical in the two algorithms.
This identity results from two points.
\begin{enumerate}
    \item No policy has $\kappa > \kappa^*$, so the first mechanism cannot be triggered;
    \item The M2V values are nonpositive (see the records for $\mathcal{M}(99)$ in Tables~\ref{tab:optProc0} and \ref{tab:optProc1}), and thus the second mechanism is prohibited.
\end{enumerate}

{\HL 
We investigate the computational efficiency of SRPI and SRPI+ under different problem sizes. 
Although Corollary~\ref{cor:complexity} states that the number of MDPs solved of SRPI is $( |\mathcal D| + 1)*( 2|\mathcal D| + 1)$, it is an upper bound for the worst case. 
We randomly generate MDPs with sizes $|\mathcal S| = |\mathcal A| \in \{ 3, 10, 30, 50, 80, 100 \}$ and run each case for 30 trials. 
Table~\ref{table:boundComparison} compares the experimental number of MDPs solved with the theoretical bound in Corollary~\ref{cor:complexity} for SRPI and SRPI+. From this table, we can see that as $|\mathcal S|$ ($|\mathcal A|$) increases, the number of MDPs solved of either SRPI or SRPI+
increases at an almost linear rate. Conversely, the upper bound, $( |\mathcal D| + 1)*( 2|\mathcal D| + 1)$, increases exponentially. This also demonstrates that SRPI and SRPI+ are actually computationally efficient in practice.
Besides, SRPI+ solves fewer MDPs than that of SRPI in 990 of 1000 trials for the case $|\mathcal S| = |\mathcal A| = 3$ (and the same numbers of MDPs solved in 3 trials), which resonates that SRPI+ is statistically faster than SRPI.

\renewcommand{\arraystretch}{1.3}
\begin{table}[H]
    \centering
        \caption{Comparisons between experimental number of MDPs solved and the theoretical bound in  Corollary~\ref{cor:complexity} for SRPI and SRPI+ with different problem sizes ($|\mathcal S| = |\mathcal A|$).} 
{ \tiny
    \begin{tabular}{ccccccc}
        \hline
        $|\mathcal S| = |\mathcal A|$ & 3  & 10  & 30  & 50  & 80  & 100 \\ \hline
        $(|\mathcal D|+1)*(2|\mathcal D|+1) \approx $ & 1540  & $2*10^{20}$  & $8*10^{88}$  & $2*10^{170}$  & $3*10^{304}$  & $2*10^{400}$ \\ \hline
        $\#$ of MDPs solved (SRPI) &  $20.70 \pm 7.04$ & $144.33 \pm 31.88$  & $900.37 \pm 204.91$  & $2113.50 \pm 314.53$  & $4365.70 \pm 888.59$  & $7007.90 \pm 1635.08$ \\ \hline
        $\#$ of MDPs solved (SRPI+) & $13.37 \pm 3.57$  & $92.67 \pm 39.33$  & $405.90 \pm 56.34$  & $949.80 \pm 148.23$  & $1997.60 \pm 119.76$  & $3140.40 \pm 422.65$ \\ \hline
    \end{tabular}\label{table:boundComparison}}
\end{table}
}

\section{Summary and outlook}\label{section5}

We present DP-based algorithms for the Sharpe ratio optimization in both average and discounted MDPs, resolving the persistent challenges caused by the fractional form of objective and the failure of DP principle.
Under Assumptions~\ref{assumption:risk_free}-\ref{assumption:square}, we reformulate the Sharpe ratio optimization to (P1) and solve it with a sequence of M2V optimizations.
The proposed resolution has a trilevel structure, where the outer level solves (P1) with a sequence of M2V optimizations, the middle level deals each M2V optimization with a sequence of MDPs, and the inner level solves standard MDPs with DP algorithms.
Taking policy iteration as the inner DP solver for example, we propose the SRPI and SRPI+ algorithms.
To the best of our knowledge, our algorithms are the first DP-based algorithms to the Sharpe ratio optimization in MDPs with a global optimality guarantee.

Our work bridges the gap between Sharpe ratio optimization and DP theory.
It provides an inspiring example on how to globally optimize fractional objectives with risk metrics in both average and discounted MDPs.
Similar algorithms could be developed for other ratio optimizations, such as Rachev ratio (with CVaR) in infinite-horizon MDPs and Sharpe ratio of accumulated rewards at a terminal stage.
Moreover, our work provides a theoretical foundation for the development of temporal-difference RL algorithms in risk-sensitive settings, which may inspire a data-driven version of our methodology.





\bibliographystyle{pomsref}

\let\oldbibliography\thebibliography
\renewcommand{\thebibliography}[1]{%
	\oldbibliography{#1}%
	\baselineskip14pt 
	\setlength{\itemsep}{10pt}
}


\end{document}